%% file: lerf.tex
\documentclass[lettersize,journal]{IEEEtran}

\usepackage{amsmath,amsfonts}
\usepackage{algorithmic}
\usepackage{algorithm}
\usepackage{array}
\usepackage[caption=false,font=normalsize,labelfont=sf,textfont=sf]{subfig}
\usepackage{textcomp}
\usepackage{stfloats}
\usepackage{url}
\usepackage{verbatim}
\usepackage{graphicx}
\usepackage{cite}
\usepackage{graphicx}
\usepackage{amsmath}
\usepackage{amssymb}
\usepackage{booktabs}
\usepackage{blindtext}
\usepackage{multirow}
\usepackage[table]{xcolor}
\usepackage{threeparttable}

\usepackage{adjustbox}

\usepackage{pifont}
\usepackage{array}

\usepackage{color, colortbl}
\usepackage{marvosym}

\newcolumntype{H}{>{\setbox0=\hbox\bgroup}c<{\egroup}@{}}

\begin{document}

\title{LeRF: Learning Resampling Function for \\ Adaptive and Efficient Image Interpolation}

\author{
        Jiacheng~Li,~
        Chang~Chen\thanks{Jiacheng~Li and Zhiwei~Xiong are with the University of Science and Technology of China, Hefei, Anhui 230026, China. E-mail: jclee@mail.ustc.edu.cn, zwxiong@ustc.edu.cn. This work is partially done during Jiacheng's internship at Noah's Ark Lab.},
        Fenglong~Song,
        Youliang~Yan\thanks{Chang~Chen, Fenglong~Song, and Youliang~Yan are with Noah's Ark Lab, Huawei Technologies. E-mail: \{chenchang25, songfenglong, yanyouliang\}@huawei.com.},
        and~Zhiwei~Xiong} 

\markboth{Journal of \LaTeX\ Class Files,~Vol.~14, No.~8, August~2021}%
{Li \MakeLowercase{\textit{et al.}}: LeRF: Learning Resampling Function for Adaptive and Efficient Image Interpolation}


\maketitle
\begin{abstract}
  \input{parts/abstract.tex}

  \end{abstract}




\input{parts/introduction.tex}
\input{parts/related.tex}
\input{parts/method.tex}

\input{parts/experiments.tex}

\input{parts/conclusion.tex}


\bibliographystyle{IEEEtran}
\bibliography{IEEEabrv,bib}

 
%

\end{document}

%% file: parts/abstract.tex
Image resampling is a basic technique that is widely employed in daily applications, such as camera photo editing.
Recent deep neural networks (DNNs) have made impressive progress in performance by introducing learned data priors.
Still, these methods are not the perfect substitute for interpolation, due to the drawbacks in efficiency and versatility.
In this work, we propose a novel method of Learning Resampling Function (termed LeRF), which takes advantage of both the structural priors learned by DNNs and the locally continuous assumption of interpolation.
Specifically, LeRF assigns spatially varying resampling functions to input image pixels and learns to predict the hyper-parameters that determine the shapes of these resampling functions with a neural network. 
Based on the formulation of LeRF, we develop a family of models, including both efficiency-orientated and performance-orientated ones.
To achieve interpolation-level efficiency, we adopt look-up tables (LUTs) to accelerate the inference of the learned neural network.
Furthermore, we design a directional ensemble strategy and edge-sensitive indexing patterns to better capture local structures.
On the other hand, to obtain DNN-level performance, we propose an extension of LeRF to enable it in cooperation with pre-trained upsampling models for cascaded resampling.
Extensive experiments show that the efficiency-orientated version of LeRF runs as fast as interpolation, generalizes well to arbitrary transformations, and outperforms interpolation significantly, \emph{e.g.}, up to 3dB PSNR gain over Bicubic for $\times2$ upsampling on Manga109. 
Besides, the performance-orientated version of LeRF reaches comparable performance with existing DNNs at much higher efficiency, \emph{e.g.}, less than 25\% running time on a desktop GPU.

%% file: parts/introduction.tex
\section{Introduction}

Due to the rapid growth of visual data, there is a strong demand for digital image processing. Image resampling is one of the most common techniques, aiming to obtain another image by generating new pixels following a geometric transformation rule from existing pixels in a given image \cite{dodgson1992image}. Common transformations include upsampling (\emph{i.e.}, single image super-resolution), downsampling, homographic transformation, etc. Image resampling enjoys various applications, ranging from photo editing, optical distortion compensation \cite{Gardner:48}, online content streaming \cite{mrak2016high}, and visual special effects production \cite{smith1986industrial}.

Recently, deep neural networks (DNNs) have made impressive progress in the field of image resampling \cite{DBLP:conf/eccv/DongLHT14,DBLP:conf/cvpr/KimLL16a,DBLP:conf/cvpr/LimSKNL17,DBLP:conf/cvpr/HuM0WT019,DBLP:journals/tip/SunC20,DBLP:conf/cvpr/SonL21}, thanks to the learning-from-data paradigm that obtains powerful structural priors from large-scale datasets. Despite the superior performance that DNN-based methods have achieved, long-lived interpolation methods like Bicubic \cite{Keys1981CubicCI} are still preferred choices in most existing devices.

We attribute this phenomenon to the following two reasons: 1) Interpolation is simple and highly efficient, resulting in less dependency and thus the practicality to be deployed on a variety of devices, ranging from IoT devices to gaming workstations. 2) Interpolation supports arbitrary transformations. It assumes a continuous resampling function for a local area, resulting in the versatility in applying to not only homographic transformations like upsampling and downsampling but also arbitrary warping. Although recent DNN-based methods explore beyond fixed-scale upsampling \cite{DBLP:conf/cvpr/LimSKNL17,DBLP:conf/cvpr/HuM0WT019, DBLP:conf/cvpr/ChenL021, DBLP:conf/cvpr/SonL21, DBLP:conf/iccv/Wang0L0AG21, DBLP:conf/nips/YangSYL21}, an efficient and continuous solution that matches interpolation remains less explored.

In this work, we aim to fill this blank research area by taking a middle way between DNN-based methods and interpolation methods. We propose a novel interpolation method of \underline{Le}arning Resampling Function (termed LeRF), where parameterized continuous functions for resampling different structures are learned from data. Specifically, as illustrated in Fig.~\ref{fig:teaser}, we assign spatially varying resampling functions to all pixels in an image, whose orientations are parameterized with several hyper-parameters. Then, we train a neural network to predict these hyper-parameters for each pixel, thus defining the resampling function for that pixel location. Finally, we obtain the resampled image by interpolating the image with these locally adapted resampling functions. LeRF takes advantage of both the structural priors learned by DNNs and the locally continuous assumption of interpolation. 

Based on the formulation of LeRF, we develop a family of models, including both efficiency-orientated and performance-orientated ones.
To achieve interpolation-level efficiency, we design an efficient implementation, where the inference of the trained neural network is accelerated with look-up tables (LUTs) \cite{DBLP:conf/cvpr/JoK21,splut,mulut,mulut23}. We further design a directional ensemble strategy and edge-sensitive indexing patterns to better capture local structures in images. On the other hand, to obtain DNN-level performance, we incorporate a pre-processing stage to enable LeRF in cooperation with pre-trained upsampling models for cascaded resampling.

We examine the advantages and generalization capacity of LeRF in various image resampling tasks, including arbitrary-scale upsampling, homographic warping, and arbitrary warping. In particular, as illustrated in Fig.~\ref{fig:tradeoff}, at a similar running time, our efficiency-orientated models (LeRF-L, LeRF-G, and LeRF-Net) outperform popular interpolation methods significantly in upsampling. Besides, the performance-orientated model (LeRF-Net++) reaches comparable performance with existing DNNs at higher efficiency. In summary, LeRF demonstrates superiority in terms of performance-efficiency tradeoff and shows versatility across different computational constraints. 

\input{figures/fig1_teaser.tex}

A preliminary version of LeRF, named LeRF-G(aussian) in this paper, appears in \cite{lerf_cvpr}, where it is dedicated to LUT-based efficient image resampling with a specific resampling function family (\emph{i.e.}, anisotropic Gaussian). In this paper, we extend LeRF substantially towards a universal paradigm for image resampling in the following aspects. 
1) We generalize the formulation of the resampling function assumption and present a more efficient resampling function family (\emph{i.e.}, amplified linear), thus obtaining a more efficient version of LeRF, \emph{i.e.}, LeRF-L(inear).
2) We design a DNN-based version of LeRF, \emph{i.e.}, LeRF-Net, achieving better performance than LeRF-G with increased receptive field (RF) and parameter capacity of a lightweight DNN.
3) We incorporate a pre-processing stage to enable LeRF in cooperation with pre-trained backbones, obtaining a performance-orientated LeRF model, \emph{i.e.}, LeRF-Net++, which matches DNN-level performance at much higher efficiency.
4) We provide a more thorough literature review, a clearer motivation of LeRF, more comprehensive experimental settings and results, as well as in-depth discussions on open questions.

Contributions of this paper are summarized as follows:

1) We propose LeRF, a novel interpolation method for image resampling. We assign spatially varying resampling functions to image pixels, where we train a neural network to predict the hyper-parameters that determine the shapes of these resampling functions.

2) We present an efficient implementation of LeRF by adopting look-up tables to accelerate the inference of the trained neural network. Furthermore, we design a directional ensemble strategy and edge-sensitive indexing patterns to better capture local structures.

3) We extend LeRF to incorporate with pre-trained models for cascaded resampling, enabling existing fixed-scale upsampling methods to support arbitrary warping.

4) Extensive experiments demonstrate that the efficiency-orientated LeRF models operate as efficiently as interpolation, generalize well to arbitrary transformations, and obtain significantly better performance over interpolation. Meanwhile, the performance-orientated LeRF model reaches comparable performance with tailored DNNs at much higher efficiency.

%% file: figures/fig1_teaser.tex
\begin{figure}[t]
    \centering
    \includegraphics[width=1\columnwidth]{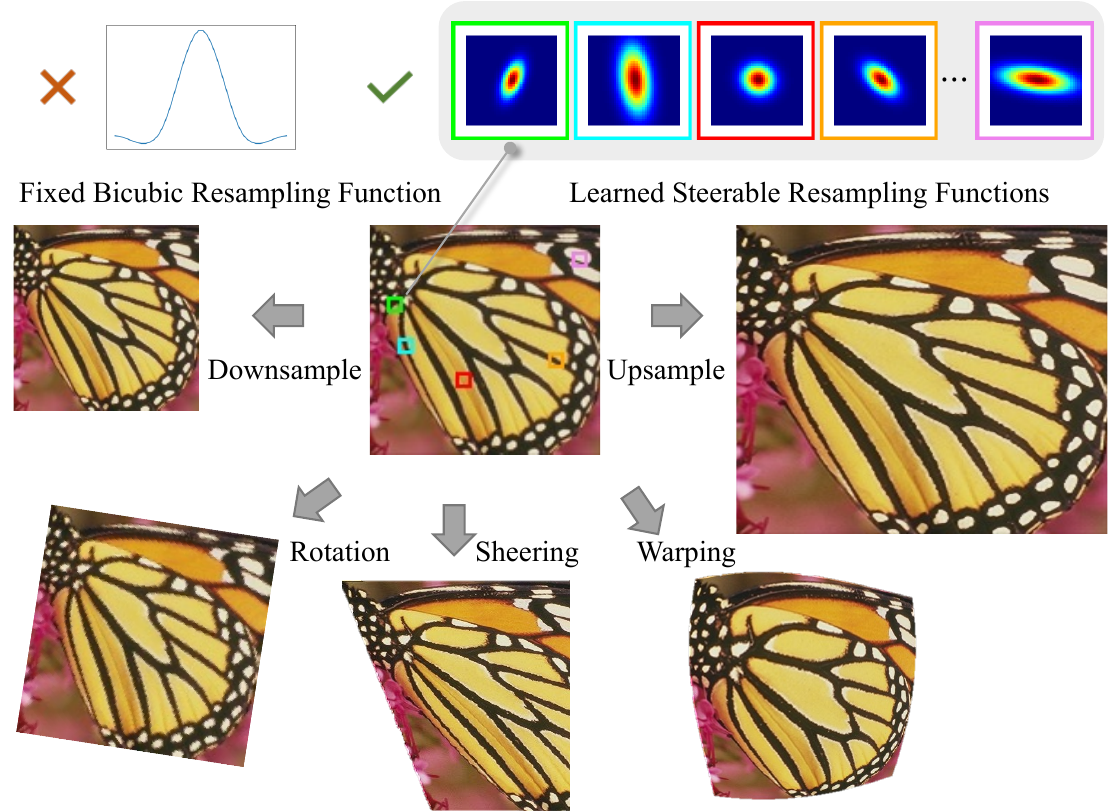}
    \caption{LeRF assigns resampling functions to input pixels and learns to predict the hyper-parameters that determine the orientations of these continuous functions for resampling under arbitrary transformations.} 
    \label{fig:teaser}
\end{figure}

%% file: parts/related.tex
\section{Related Works}

\noindent\textbf{Image interpolation for resampling.}
Interpolation, the most common solution for image resampling, assumes a locally continuous intensity surface and approximates it with a fixed resampling function, such as $\mathtt{nearest}$ (Nearest), $\mathtt{linear}$ (Bilinear), $\mathtt{cubic}$ (Bicubic) \cite{Keys1981CubicCI}, and windowed $\mathtt{sinc}$ (Lanczos). It predicts resampling weights with the assumed resampling function, and then aggregates input pixels with these resampling weights to obtain the target pixel. This assumption allows for continuous resampling under arbitrary transformations, yet leads to blurry results due to ignoring different local structures \cite{DBLP:conf/siggraph/MitchellN88}. We follow the same assumption on local continuity, but our method deals with different structures with adapted resampling functions instead of fixed ones.

\input{figures/fig_tradeoff_gpu.tex}

\noindent\textbf{Adaptive image resampling.}
To integrate local structural information into the resampling process, many adaptive interpolation methods are proposed, including edge-directed interpolation \cite{DBLP:conf/icip/AllebachW96, DBLP:journals/tip/LiO01a, DBLP:journals/tip/ZhangW06a, DBLP:journals/tip/WangW07}, \emph{e.g.}, NEDI \cite{DBLP:journals/tip/LiO01a}, and kernel regression \cite{DBLP:journals/tip/TakedaFM07, DBLP:journals/tip/LeeY10, DBLP:journals/tip/ZhangGTL12}, \emph{e.g.}, SKR \cite{DBLP:journals/tip/TakedaFM07}. Different from these methods that rely on hand-designed rules, our method adopts a neural network to learn structural priors and integrate them into the resampling functions in a data-driven way.
Another line of work achieves adaptive resampling by combining interpolation methods with adaptive filtering \cite{DBLP:conf/cvpr/KimLL16a,DBLP:journals/tci/RomanoIM17,DBLP:conf/iccp/GetreuerGICOM18,DBLP:journals/corr/abs-1712-06463}. Among them, RAISR achieves super-resolution by predicting adaptive filters for each pixel from a hash table and then applying the predicted adaptive filters to pre-upsampled images \cite{DBLP:journals/tci/RomanoIM17, DBLP:conf/iccp/GetreuerGICOM18}. This kind of method operates on the pre-resampled results of common interpolation methods (usually Bicubic) with a fixed transformation (\emph{e.g.}, $\times2$ upsampling), thus lacking the generalization ability to be extended to unseen arbitrary transformations.

\noindent\textbf{DNN-based image resampling.}
With the rise of deep neural networks, impressive progress has been made in the field of image transformation, such as fixed-scale upsampling \cite{DBLP:conf/eccv/DongLHT14,DBLP:conf/cvpr/KimLL16a,DBLP:conf/cvpr/LimSKNL17,DBLP:conf/iccvw/ChenTWX17,DBLP:conf/eccv/ZhangLLWZF18,DBLP:conf/cvpr/ChenXTZW19,DBLP:journals/tog/WronskiGEKKLLM19,DBLP:conf/iccvw/LiangCSZGT21,DBLP:conf/cvpr/XiaoFHCX21,pan2022towards,yao2023bidirectional} and downsampling \cite{DBLP:journals/tip/SunC20,DBLP:conf/iccv/TalebiM21}. To achieve arbitrary-scale upsampling, Meta-SR \cite{DBLP:conf/cvpr/HuM0WT019} predicts the resampling weights from upsampling scales with a weight prediction network, similar to kernel prediction networks \cite{DBLP:conf/cvpr/MildenhallBCSNC18,DBLP:conf/cvpr/Liang0GGT21}, and aggregates deep features to get the final upsampled image. Following works improve the process with scale-aware feature adaptation \cite{DBLP:conf/iccv/Wang0L0AG21}, scale-encoded feature fusion \cite{DBLP:conf/cvpr/0004CNWTL23}, integration of spatially variant degradation map \cite{DBLP:conf/cvpr/BernasconiDSGS23}, and better continuous space representations \cite{DBLP:conf/cvpr/WeiZ23,DBLP:conf/cvpr/SongSZSS023}. SRwarp explores homographic warping through a more flexible kernel prediction design and a multi-scale blending strategy \cite{DBLP:conf/cvpr/SonL21}. Along the other line, arbitrary-scale upsampling methods based on implicit neural function, \emph{e.g.}, LIIF \cite{DBLP:conf/cvpr/ChenL021}, model the aggregation step with a neural implicit representation, predicting the target pixels from deep latent features and target coordinates. Following methods improve the implicit representation by integrating frequency analysis \cite{DBLP:conf/cvpr/LeeJ22}, modulation-based transformer \cite{DBLP:conf/nips/YangSYL21}, implicit attention mechanism \cite{DBLP:conf/cvpr/Cao0XLNP0ZTG23}, local implicit transformer \cite{DBLP:conf/cvpr/ChenXHTKL23}, conditional normalizing flow \cite{DBLP:conf/cvpr/YaoTLTCL23}, and implicit diffusion model \cite{DBLP:conf/cvpr/GaoLZXLLLZ023}. LTEW achieves homographic warping by incorporating coordination transformation with frequency representation into the implicit modeling process \cite{DBLP:conf/eccv/LeeCJ22}. MFR further enhance the frequency representation by introducing learnable Gabor wavelet filters \cite{Xiao_2024_CVPR}.
Different from these works that rely on kernel prediction networks or implicit neural functions, we utilize explicit local functions, resulting in advantages in efficiency and interpretability. Besides, while a few works share a similar idea of estimating hyper-parameters for explicit functions, \emph{e.g.}, b-spline \cite{DBLP:conf/cvpr/PakLJ23} and filter kernel \cite{DBLP:journals/corr/abs-2112-09318}, our work features an efficient design that matches interpolation.

\noindent\textbf{Efficient image processing with look-up tables.}
A look-up table is composed of index-value pairs, which can be efficiently retrieved through memory access. It is widely applied in the image signal processing pipeline \cite{DBLP:journals/pami/KimLLSLB12,DBLP:journals/pami/ZengCLCZ22}. Recently, SR-LUT has been introduced to accelerate the inference of a fixed-scale super-resolution network by traversing all possible low-resolution (LR) patches, pre-computing all corresponding high-resolution (HR) patches, and saving them as index-value pairs \cite{DBLP:conf/cvpr/JoK21}. At inference, the computations in the super-resolution network are replaced with retrieving values from LUT, leading to inference acceleration. Following works extend this idea with parallel design \cite{splut}, reconstructed convolutional operation \cite{DBLP:conf/iccv/LiuDLSWW23}, and DNN-like combination \cite{mulut, mulut23}. Different from existing LUT-based methods \cite{DBLP:conf/cvpr/JoK21,splut,mulut,DBLP:conf/iccv/LiuDLSWW23,mulut23}, which adopt different groups of LUTs under different upsampling scales, our LUTs store the same group of hyper-parameters across arbitrary transformations, thus achieving continuous resampling at high efficiency.

%% file: figures/fig_tradeoff_gpu.tex
\begin{figure}[t]
    \centering
    \includegraphics[width=\columnwidth]{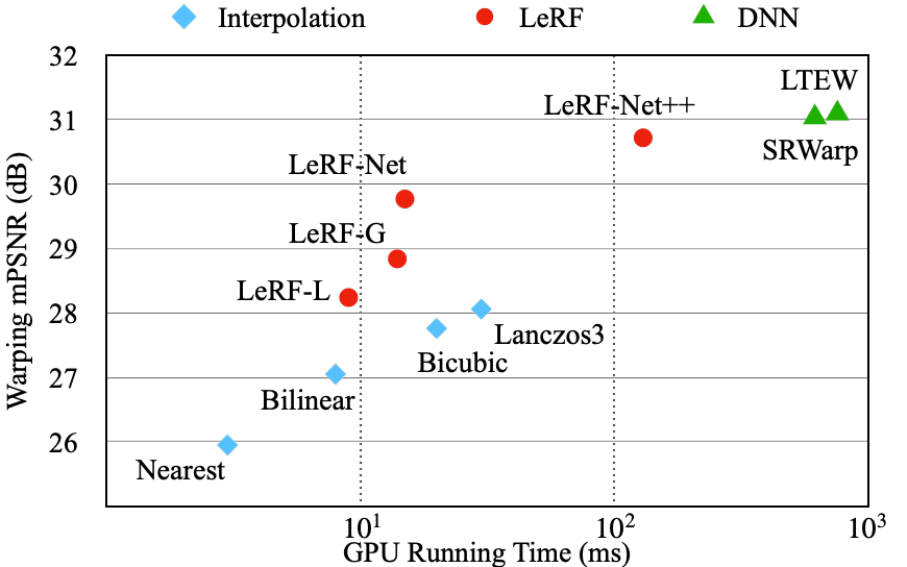}
    \caption{Performance-efficiency trade-off of image resampling methods. mPSNR values are obtained on DIV2K for in-scale homographic warping. The running time is evaluated on an NVIDIA RTX 3090 GPU for producing $1280 \times 720$ images through $\times 4$ upsampling.} 
    \label{fig:tradeoff}
\end{figure}

%% file: parts/method.tex
\section{Learning Resampling Function}

\subsection{Formulation}

Typically, as illustrated in Fig.~\ref{fig:overview}(a), image resampling through interpolation can be implemented in the following steps. 

\ding{172} \textit{Obtain relative offsets}: the target coordinates after transformation, such as upsampling, are projected back to the coordinate space of the source image, and the relative spatial offsets between target pixels and source pixels in their support patches are obtained. 

\ding{173} \textit{Predict resampling weights}: based on the relative spatial offsets, the resampling weights, \emph{i.e.}, resampling kernels, are predicted for each pixel in the support patch. 

\ding{174} \textit{Aggregate pixels}: the source pixels are aggregated through weighted summation to obtain the target pixel value. The above process can be formulated as
\begin{equation} \label{eq:weight_sum}
    \hat{I}^{q} = \sum_{p \in \Omega} W_{p \rightarrow q} I^{p}, \quad \hat{I}^{q} \in I_S, \quad I^{p} \in I_T ,
\end{equation}
where $\hat{I}^{q}$ is the interpolated pixel value at the target coordinate $q$, $I^p$ the pixel value at the source coordinate, $\Omega$ the support patch, $W_{p \rightarrow q}$ the weight from $p$ to $q$, $I_{S}$ the source image, and $I_T$ the target image.

\input{figures/fig2_overview.tex}

For interpolation methods, an isotropic resampling function $\Phi(\cdot)$, \emph{e.g.} $\mathtt{cubic}$ in Bicubic interpolation, is assumed to predict resampling weights from relative offsets, which can be formulated as
\begin{equation} \label{eq:resample}
    W_{p \rightarrow q} = \Phi( D_{p \rightarrow q}),
\end{equation}
where $D_{p \rightarrow q}$ is the relative offset from $p$ to $q$, which can be obtained based on the geometric transformation $\mathcal{T}$. Thanks to the continuity of the resampling function $\Phi$, arbitrary transformations from $I_{S}$ to $I_{T}$ can be achieved. The assumption of local continuity contributes to the versatility and high efficiency of interpolation.

Summarizing Eq.~\ref{eq:weight_sum} and Eq.~\ref{eq:resample}, we formulate interpolation methods as 
\begin{equation}
    I_T = \Phi (D) \times I_S.
\end{equation}

For adaptive resampling methods, as illustrated in Fig.~\ref{fig:overview}(b), we formulate them as
\begin{equation}
    I_T = \Gamma \otimes (\Phi (D) \times I_S),
\end{equation}
where $\Gamma$ denotes filters based on hand-designed rules, and $\otimes$ stands for the filtering process.

For DNN-based resampling methods, as shown in Fig.~\ref{fig:overview}(c), we formulate them as
\begin{equation}
    I_T = f_r(f_e(I_S), f_k(D)),
\end{equation}
where, $f_e(\cdot)$, $f_k(\cdot)$, and $f_r(\cdot)$ are feature extraction network, kernel prediction network, and reconstruction network, respectively. Methods based on implicit neural function unify $f_k(\cdot)$ and $f_r(\cdot)$, omitting the estimation of $W$. 

In this work, we propose a novel method of learning resampling function that adapts the resampling functions to local structures in a data-driven way. Different from the \emph{fixed} resampling function in interpolation, we assume a kind of \emph{adaptive} resampling function $\Phi_{\Theta}$, parameterized by $\Theta$. As illustrated in Fig.~\ref{fig:overview}(d), the proposed LeRF method can be described as
\begin{equation}
    I_T = \Phi_{\Theta} \times g(I_S), \quad \Theta = f(I_S),
\end{equation}
where $f(\cdot)$ is a DNN, $g(\cdot)$ denotes a DNN or a filter for pre-processing the source image. Next, we will discuss the choice of the adaptive resampling function $\Phi_{\Theta}(\cdot)$.

\input{figures/fig_kernel_vis.tex}

\subsection{Adaptive Resampling Function}

As illustrated in the left part of Fig.~\ref{fig:kernel_vis}, a resampling function maps coordinate offset $D$ to resampling weight $W$. 

For example, the 1D case of the $\mathtt{cubic}$ function in Bicubic interpolation is defined as 
\begin{equation} \label{eq:cubic}
    \Phi(x)= \begin{cases}(a+2)|x|^3-(a+3)|x|^2+1 & \text { for }|x| \leq 1, \\ 
        a|x|^3-5 a|x|^2+8 a|x|-4 a & \text { for } 1<|x|<2, 
        \\ 0 & \text { otherwise, }\end{cases}
\end{equation}
where $a$ is a hyper-parameter that is recommended to be set to $0.5$ in the widely applied Keys Bicubic interpolation \cite{Keys1981CubicCI}. Other formulations and hyper-parameter choices for $\mathtt{cubic}$ function are also discussed in prior work \cite{DBLP:conf/siggraph/MitchellN88}. Some image processing software like Adobe PhotoShop\footnote{https://www.adobe.com/products/photoshop.html} and Imagick\footnote{https://imagemagick.org/} support for changing this hyper-parameter by users for each image, yet lacking the flexibility across different pixels. This spatially invariant assumption for resampling function in classic interpolation methods leads to blurry results due to the ignorance of local structures \cite{DBLP:conf/siggraph/MitchellN88}.

Generally, in the proposed LeRF, for each pixel $\hat{I}^q$ at the target location $q$, the aggregation process with adaptive resampling function can be written as
\begin{equation} \label{eq:lerf}
        \hat{I}^{q} = \sum_{p \in \Omega} \Phi_{\Theta^p}(D_{p \rightarrow q}) I^p,
\end{equation}
where $\Theta^p$ is a set of data-dependent hyper-parameters of source pixel $I^p$.

There are various choices for the resampling function family to be learned. Generally speaking, it should be 1) parameterized with fewer parameters to lower learning complexity and 2) easily adapted to various local structures. Here, we introduce two examples of adaptive resampling function families, \emph{i.e.}, amplified linear and anisotropic Gaussian. The former is a generalized formulation of linear function and the latter is a steerable version of the Gaussian kernel. More complicated function families can also be integrated into LeRF.

\textit{The amplified linear resampling function.}
1D linear resampling function, shown in Fig.~\ref{fig:kernel_vis}, can be formulated as
\begin{equation}
    \Phi(x)= \begin{cases} 1 - |x| , & \text { for } 0 \leq |x| \leq 1, \\ 
         0 & \text { otherwise, }\end{cases}
\end{equation}
We generalize it by amplifying the input variable with a scalar $\alpha$, obtaining
\begin{equation}
    \Phi^{L}_{\alpha}(x)= \begin{cases} 1 - \alpha|x| , & \text { for } 0 \leq |x| \leq 1, \\ 
        0 & \text { otherwise, }\end{cases}
\end{equation}
where $\alpha > 0$. When $\alpha=1$, it degrades to the default linear resampling function. In the above amplified linear function, the hyper-parameter $\Theta$ is $\alpha$ itself. As illustrated in the right part of Fig.~\ref{fig:kernel_vis}, a larger $\alpha$ produces a more compact kernel shape, thus reducing the averaging effect between neighboring pixels. In practice, we set a maximum bound for $\alpha$ to avoid explosive gradients during optimization.

\textit{The anisotropic Gaussian resampling function.}
To empower the adaptive resampling function to steer itself along local structures, we also present a formulation with anisotropic Gaussian,
\begin{equation}
    \Phi^{G}_{{\boldsymbol  \Sigma }}(\boldsymbol {x }) = \frac{1}{2\pi{|{\boldsymbol  \Sigma }|}^{\frac{1}{2}}} exp\{-\frac{1}{2} {(\boldsymbol {x })}^T {\boldsymbol  \Sigma }^{-1} {(\boldsymbol {x })}\}
\end{equation}
where $|\cdot|$ denotes the determinant of the covariance matrix ${\boldsymbol  \Sigma }$. We parameterize ${\boldsymbol  \Sigma }$ as the following,
\begin{equation}
    {\boldsymbol  \Sigma }={\begin{pmatrix}\sigma _{X}^{2}&\rho \sigma _{X}\sigma _{Y}\\\rho \sigma _{X}\sigma _{Y}&\sigma _{Y}^{2}\end{pmatrix}} 
\end{equation}
where $\rho$, $\sigma_{X}$, and $\sigma_{Y}$ are hyper-parameters, and thus the resampling function becomes $\Phi^G_{(\rho, \sigma_X, \sigma_Y)}$. The above formulation is the same as the probability density function of bivariate normal distribution, where $\rho$ can be interpreted as the correlation between 2D variables and $\sigma_X, \sigma_Y$ the standard deviations. This formulation of anisotropic Gaussian is widely applied in the literature, including image filtering with kernel regression \cite{DBLP:journals/tip/TakedaFM07} and computer graphics \cite{DBLP:journals/tog/YuT13}.

As illustrated in the right part of Fig.~\ref{fig:kernel_vis}, the anisotropic Gaussian function is steerable, where different orientations and shapes can be obtained by tuning hyper-parameters $(\rho, \sigma_X, \sigma_Y)$, showing its modeling capacity for a variety of local structures. 
But, direct optimization of the above hyper-parameters leads to unstable gradients and divergence. Thus, we modify the formulation by omitting the determinant multiplier and predicting a group of $(\rho, \frac{1}{\sigma_{X}}, \frac{1}{\sigma_{Y}})$ for stable training. In the following section, we describe the learning process of LeRF.

\input{figures/fig_pattern.tex}

\subsection{Hyper-parameter Learning and Pre-processing}

As shown in Fig.~\ref{fig:overview}(d), we propose to learn the estimation of hyper-parameters in the adaptive resampling function in a data-driven way. Specifically, we adopt a hyper-parameter learning network $f(\cdot)$ to predict $\Theta$ from the source image $I_S$. For an image with shape $H \times W$, $f(\cdot)$ outputs a tensor of $\Theta$ with a size of $H \times W \times C$. We set $C=1$ for the amplified linear resampling function and $C=3$ for the anisotropic Gaussian resampling function. That is to say, LeRF predicts a set of hyper-parameters for each source pixel.

Inspired by DNN-based resampling methods \cite{DBLP:conf/cvpr/HuM0WT019,DBLP:conf/iccv/Wang0L0AG21,DBLP:conf/nips/YangSYL21}, we pre-process the image with an additional neural network $g(\cdot)$. Pre-processing network $g(\cdot)$ can be viewed as a feature extractor to enhance object structures. Furthermore, as shown later in Fig.~\ref{fig:visual_abl_pre}, this pre-processing stage is configurable. It can be a pre-trained fixed-scale upsampler to pre-upsample the image for richer high-frequency details (see Sec.~\ref{sec:lerfnet_plus}), or serve as a low-pass filter to alleviate the aliasing artifacts when downsampling (see Fig.~\ref{fig:visual_down}).

Thanks to the differentiability of the adaptive resampling function, LeRF can be trained on external data pairs with the Mean-Squared Error (MSE) loss function in an end-to-end manner. The training loss function at target coordinate $q$ can be formulated as
\begin{equation} \label{eq:training}
    \mathcal{L}^q = || I^q_T - \sum_{p \in \Omega} \Phi_{\Theta}(D_{p \rightarrow q}) g(I^p_S) ||^2, \quad \Theta = f(\mathcal{N}_p),
  \end{equation}
where $I^q_T$ is the ground truth pixel value at location $q$ and $\mathcal{N}_p$ denotes the surrounding pixels of $I_p$ (not necessarily the same as the support patch $\Omega$). In practice, we utilize fixed-scale upsampling as a proxy task and train LeRF on LR and HR image pairs.

\section{Application}

In this section, we show the family of models based on LeRF, including both efficiency-orientated and performance-orientated ones. Different configurations of the LeRF model are summarized in TABLE~\ref{tab:config}.

\subsection{Efficiency-orientated LeRF Model}

To match the efficiency of interpolation, we accelerate the inference of the trained DNNs, \emph{i.e.}, $f(\cdot)$ and $g(\cdot)$, by adopting look-up tables \cite{DBLP:conf/cvpr/JoK21,splut,mulut,mulut23}. As shown in Fig.~\ref{fig:pattern}(a), for each pair in the LUT to accelerate DNN, its index $\boldsymbol i_{*}$ is a combination of pixels (\emph{i.e.}, $\mathcal{N}_p$), and its value $\boldsymbol v_{*}$ is a group of corresponding hyper-parameters for that pixel combination (\emph{i.e.}, $\Theta$). This way, hyper-parameters can be retrieved directly from the saved values in LUTs, skipping computations in DNN $f(\cdot)$ and thus resulting in high efficiency. The pre-precessing network $g(\cdot)$ can also be accelerated by LUTs.

\input{tables/config_table.tex}

Different from existing LUT-based methods, whose LUT values are image pixels, our LUTs store hyper-parameters that reflect structural characteristics. Thus, to better extract structural priors, we propose the following adaptations.

\textit{Directional ensemble strategy.} We propose a directional ensemble (DE) strategy to replace the rotation ensemble (RE) strategy in existing LUT-based methods \cite{DBLP:conf/cvpr/JoK21,mulut,mulut23}. As illustrated in Fig.~\ref{fig:pattern}(b), in RE, the predictions are averaged across all directions, while the proposed DE strategy only ensembles the predictions with the same direction (\emph{i.e.}, $180^{\circ}$ rotational symmetry, instead of $90^{\circ}$ in RE). This enables the learning of $\rho$, which determines the orientation of the steerable resampling function (see TABLE~\ref{tab:abl3} and Fig.~\ref{fig:visual_abl2}).

\input{figures/fig_net_arch.tex}

\textit{Edge-sensitive indexing patterns.} As illustrated in Fig.~\ref{fig:pattern}(c), we include patterns ``$C$'' and ``$X$'', alongside the default ``$S$'' pattern in SR-LUT \cite{DBLP:conf/cvpr/JoK21} to better capture edges of different orientations. For example, The ``$C$'' and ``$C'$'' patterns are sensitive to vertical and horizontal edges, respectively. We validate their effectiveness in TABLE~\ref{tab:abl3}. Correspondingly, as shown in Fig.~\ref{fig:pattern}(a), the hyper-parameter learning network follows a multi-branch design, and each branch is accelerated by a LUT. Branches in this network share the same architecture, which is shown in Fig.~\ref{fig:net_arch}.

With the above acceleration implementation, we obtain two configurations of LeRF models, \emph{i.e.}, LeRF-L(inear) and LeRF-G(aussian). Compared to LeRF-G which appears in our previous work \cite{lerf_cvpr}, LeRF-L shares the same network architecture but differs in the adaptive resampling used. LeRF-L utilizes the amplified linear resampling function while LeRF-G the anisotropic Gaussian. As discussed in previous LUT-based image processing methods \cite{mulut, mulut23}, the network architecture design is limited due to the limitation in the receptive field of LUT. Here, inspired by IMDN \cite{DBLP:conf/mm/HuiGYW19}, we modify a lite version of IMDN to further enhance the ability of the hyper-parameter learning network and pre-processing network with increased receptive field and parameter capacity, as illustrated in Fig.~\ref{fig:net_arch}. We name this version of LeRF as LeRF-Net. As shown later in TABLE~\ref{tab:comp}, although LeRF-Net is not as efficient as LeRF-G in resource-limited devices (\emph{e.g.}, mobile CPU), it shows comparable efficiency on GPU and superior performance.

\input{tables/main_table.tex}

\input{figures/visual_upsample.tex}

\subsection{Performance-orientated LeRF model} \label{sec:lerfnet_plus}


To obtain DNN-level performance, we propose to adapt LeRF to pre-upsampled results with pre-trained fixed-scale upsampling models. Specifically, we adopt an RCAN \cite{DBLP:conf/eccv/ZhangLLWZF18} model as the pre-processing network $g(\cdot)$, pre-upsampling the image into two times the original size. Then we fine-tune LeRF-Net on these pre-processing results and obtain a performance-orientated LeRF model, LeRF-Net++. This adaptation process enables LeRF in cooperation with pre-trained models for cascaded resampling, showing the flexibility of LeRF. As shown later in Fig.~\ref{fig:tradeoff_cpu}, LeRF-Net++ reaches comparable performance with existing DNNs at higher efficiency.

%% file: figures/fig2_overview.tex
\begin{figure*}[t]
    \centering
    \includegraphics[width=0.9\textwidth]{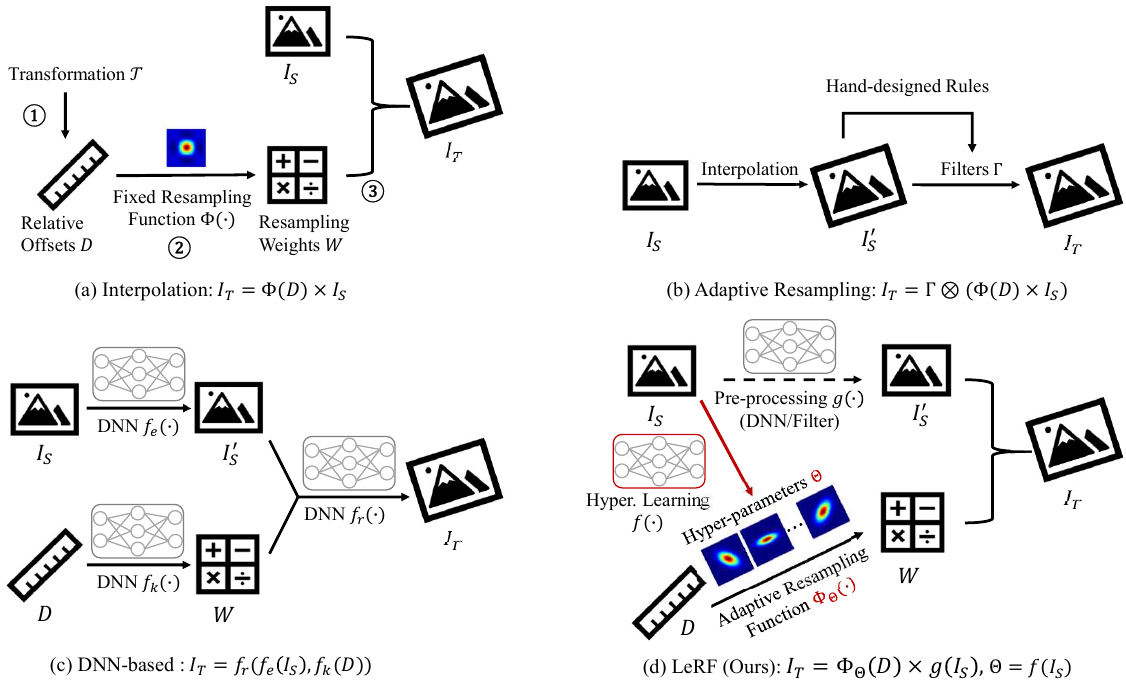}
    \caption{Comparison of exiting resampling methods and LeRF. 
    (a) Interpolation assumes a spatial-invariant fixed function $\Phi(\cdot)$ to predict resampling weights $W$. Representative method: Bilinear, Bicubic.
    (b) adaptive resampling methods interpolate or filter the interpolated image based on hand-designed rules $\Gamma$. Representative method: NEDI \cite{DBLP:journals/tip/LiO01a} and RAISR \cite{DBLP:journals/tci/RomanoIM17}.
    (c) DNN-based methods rely on feature extraction network ($f_e(\cdot)$), reconstruction network ($f_r(\cdot)$), and kernel prediction network ($f_{k}(\cdot)$) or implicit neural function (unified $f_{k}(\cdot)$ and $f_{r}(\cdot)$). Representative method: Meta-SR \cite{DBLP:conf/cvpr/HuM0WT019}, LIIF \cite{DBLP:conf/cvpr/ChenL021}, SRwarp \cite{DBLP:conf/cvpr/SonL21}, and LTEW \cite{DBLP:conf/eccv/LeeCJ22}.
    (b) LeRF learns spatially varying resampling function $\Phi_{\Theta}$ with a DNN $f(\cdot)$ that predicts hyper-parameters $\Theta$ for each pixel.} 
    \label{fig:overview}
\end{figure*}

%% file: figures/fig_kernel_vis.tex
\begin{figure*}[t]
    \centering
    \includegraphics[width=0.9\textwidth]{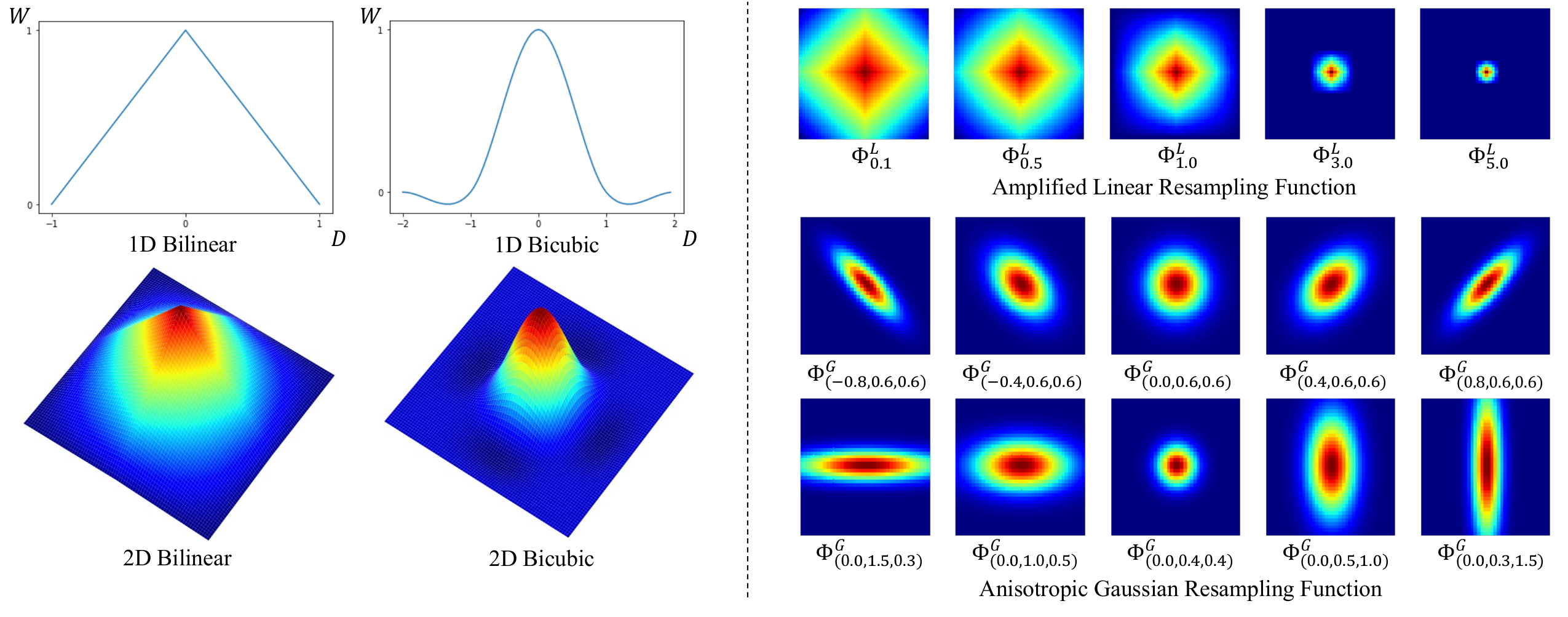}
    \caption{Visualization of fixed resampling functions (left) and adaptive resampling functions (right).} 
    \label{fig:kernel_vis}
\end{figure*}

%% file: figures/fig_pattern.tex
\begin{figure*}[t]
    \centering
    \includegraphics[width=0.9\textwidth]{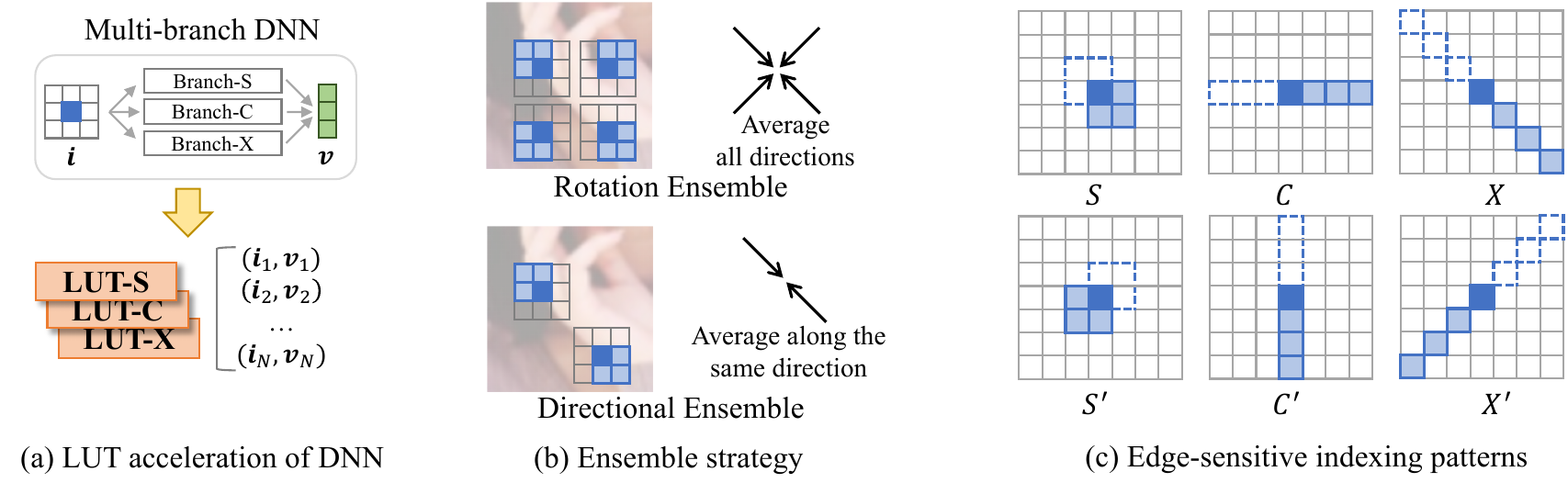}
    \caption{LUT acceleration of DNN and our adaptations. (a) A learned neural network can be accelerated by traversing all possible inputs $i_*$, pre-computing all corresponding outputs $v_*$, and saving them to LUTs \cite{DBLP:conf/cvpr/JoK21,mulut,mulut23,splut}. (b) In the proposed directional ensemble strategy, only predictions along the same direction are averaged, instead of all directions in rotation ensemble \cite{DBLP:conf/cvpr/JoK21}. (c) We introduce edge-sensitive indexing patterns to better capture edge orientations. The pixels covered by directional ensemble are depicted with dashed boxes.} 
    \label{fig:pattern}
\end{figure*}

%% file: tables/config_table.tex
\newcommand{\specialcell}[2][c]{%
  \begin{tabular}[#1]{@{}r@{}}#2\end{tabular}}
  
\renewcommand{\specialcell}[2][l]{%
  \begin{tabular}[#1]{@{}l@{}}#2\end{tabular}}

\begin{table}[t]
    \caption{Different configurations of the LeRF model family. We name the model in our previous work \cite{lerf_cvpr} as LeRF-G in this paper.}
    \label{tab:config}
    \centering
    \resizebox{\columnwidth}{!}{%
    \begin{tabular}{llll}
    \toprule
    \multicolumn{4}{c}{Efficiency-orientated LeRF Models} \\
    \midrule
     Config. & $g(\cdot)$ & $f(\cdot)$ & $\Phi_{\Theta}(\cdot)$ \\
     \midrule
    LeRF-L       & LUT-\{$S$,$C$,$X$\}    & LUT-\{$S$,$C$,$X$,$S'$,$C'$,$X'$\}    & Linear      \\
    LeRF-G \cite{lerf_cvpr}     & LUT-\{$S$,$C$,$X$\}    & LUT-\{$S$,$C$,$X$,$S'$,$C'$,$X'$\}   & Gaussian   \\
    LeRF-Net       & lightweight DNN    & lightweight DNN   &  Gaussian   \\
    \midrule \midrule
    \multicolumn{4}{c}{Performance-orientated LeRF Model} \\
    \midrule
    Config. & $g(\cdot)$ & $f(\cdot)$ & $\Phi_{\Theta}(\cdot)$ \\
    \midrule
    LeRF-Net++       & RCAN ($\times 2$) \cite{DBLP:conf/eccv/ZhangLLWZF18}    & lightweight DNN   & Gaussian   \\
    \bottomrule
    \end{tabular}
    }
  \end{table}

  \renewcommand{\specialcell}[2][c]{%
  \begin{tabular}[#1]{@{}r@{}}#2\end{tabular}}

%% file: figures/fig_net_arch.tex
\begin{figure}[t]
    \centering
    \includegraphics[width=\columnwidth]{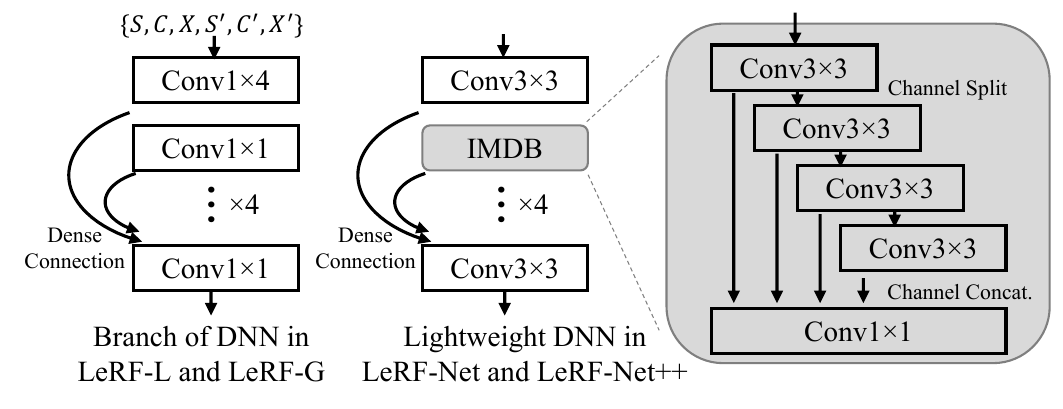}
    \caption{Detailed network architectures of multi-branch DNN for LUT acceleration (LeRF-L and LeRF-G) and the IMDN-like lightweight DNN (LeRF-Net and LeRF-Net++).} 
    \label{fig:net_arch}
\end{figure}

%% file: tables/main_table.tex
\newcolumntype{C}{>{\centering\arraybackslash}p{0.035\textwidth}}
\definecolor{Gray}{gray}{0.95}
\renewcommand{\arraystretch}{1.2}

\begin{table*}[t]
    \caption{Quantitative comparison in PSNR for arbitrary-scale upsampling.}
    \begin{threeparttable}
    \resizebox{\textwidth}{!}{%
    \begin{tabular}{lCCCCCCCCCCCCCCCCCCCC}
        
    \toprule
    \multirow{2}{*}{Method} &\multicolumn{4}{c}{Set5} &\multicolumn{4}{c}{Set14} &\multicolumn{4}{c}{BSDS100} &\multicolumn{4}{c}{Urban100} &\multicolumn{4}{c}{Manga109} \\ \cmidrule(lr){2-5} \cmidrule(lr){6-9} \cmidrule(lr){10-13} \cmidrule(lr){14-17} \cmidrule(lr){18-21} 
                           &$\frac{\times1.5}{\times1.5}$   &$\frac{\times1.5}{\times2.0}$   &$\frac{\times2.0}{\times2.0}$   &$\frac{\times2.0}{\times2.4}$
                        &$\frac{\times1.5}{\times1.5}$   &$\frac{\times1.5}{\times2.0}$   &$\frac{\times2.0}{\times2.0}$   &$\frac{\times2.0}{\times2.4}$
                        &$\frac{\times1.5}{\times1.5}$   &$\frac{\times1.5}{\times2.0}$   &$\frac{\times2.0}{\times2.0}$   &$\frac{\times2.0}{\times2.4}$
                        &$\frac{\times1.5}{\times1.5}$   &$\frac{\times1.5}{\times2.0}$   &$\frac{\times2.0}{\times2.0}$   &$\frac{\times2.0}{\times2.4}$
                        &$\frac{\times1.5}{\times1.5}$   &$\frac{\times1.5}{\times2.0}$   &$\frac{\times2.0}{\times2.0}$   &$\frac{\times2.0}{\times2.4}$ \\ \midrule
                                    Nearest  &31.34	&31.07	&30.84	&29.63   
                                            &29.15	&28.84	&28.57	&27.70  
                                            &28.99	&28.72	&28.40	&27.62 
                                            &26.21	&25.91	&25.62	&24.78
                                            &28.59	&28.36	&28.14	&26.87  
                                    \\
                                   Bilinear   &34.99	&33.19	&32.23	&31.49
                                            &31.68	&30.26	&29.24	&28.70
                                            &30.92	&29.66	&28.67	&28.20
                                            &28.24	&26.91	&25.96	&25.46
                                            &32.45	&30.33	&29.16	&28.28
                                    \\
                                   Bicubic    &36.76	&34.68	&33.64	&32.70
                                            &33.07	&31.45	&30.32	&29.62
                                            &32.14	&30.67	&29.54	&28.93
                                            &29.50	&27.95	&26.87	&26.22
                                            &34.76	&32.13	&30.81	&29.61   
                                    \\
                                   Lanczos2  &36.83	&34.74	&33.70	&32.74
                                            &33.13	&31.50	&30.36	&29.65
                                            &32.19	&30.71	&29.58	&28.95
                                            &29.55	&28.00	&26.91	&26.25
                                            &34.87	&32.22	&30.89	&29.66
                                     \\  
                                   Lanczos3  &37.61	&35.31	&34.23	&33.24
                                            &{33.75}	&{31.97}	&30.76	&30.02
                                            &32.74	&31.11	&29.89	&29.23
                                            &30.12	&28.42	&27.25	&26.55
                                            &{36.12}	&33.06	&{31.63}	&30.28
                                     \\ 
                                    RAISR* \cite{DBLP:journals/tci/RomanoIM17}      &35.50	&35.49	&{35.57}	&33.38   
                                           &31.84	&31.67	&{31.71}	&{30.22}  
                                           &30.87	&30.68	&{30.66}	&29.42 
                                           &28.77	&{28.60}	&{28.64}	&{27.01}
                                           &33.81	&{33.74}	&\underline{33.88}	&{30.61}  
                                           \\
                                    LeRF-L        &37.77	&35.84	&34.84	&33.60   
                                            &33.99	&32.40	&31.22	&30.32  
                                            &33.24	&31.59	&30.32	&29.56 
                                            &30.53	&28.84	&27.65	&26.85
                                            &36.42	&33.75	&32.35	&30.78  
                                            \\ 
                                    LeRF-G        &\underline{38.30}	&\underline{36.60}	&\underline{35.71}	&\underline{34.74}   
                                    &\underline{34.59}	&\underline{33.06}	&\underline{31.98}	&\underline{31.10}  
                                    &\underline{33.76}	&\underline{32.08}	&\underline{30.83}	&\underline{30.09} 
                                    &\underline{31.86}	&\underline{30.08}	&\underline{28.86}	&\underline{27.99}
                                    &\underline{36.57}	&\underline{34.79}	&\underline{33.88}	&\underline{32.67}  
                                    \\  

                                    LeRF-Net        &\textbf{38.98}	&\textbf{37.36}	&\textbf{36.57}	&\textbf{35.59}   
                                            &\textbf{35.44}	&\textbf{33.78}	&\textbf{32.68}	&\textbf{31.81}  
                                            &\textbf{34.47}	&\textbf{32.80}	&\textbf{31.54}	&\textbf{30.75} 
                                            &\textbf{32.91}	&\textbf{31.34}	&\textbf{30.29}	&\textbf{29.36}
                                            &\textbf{38.06}	&\textbf{36.25}	&\textbf{35.37}	&\textbf{34.21}  
                                            \\  
                                            
                                            \midrule
RCAN* \cite{DBLP:conf/eccv/ZhangLLWZF18}                      &41.16	&\underline{39.26}	&\textbf{38.27}	&36.91   
&37.19  	&\underline{35.52}	&\textbf{34.12}	&32.84  
&35.64  	&\underline{33.82}	&\textbf{32.41}	&31.42
&36.06  	&\underline{34.46}	&\textbf{33.35}	&31.37
&42.53  	&\underline{40.65}	&\textbf{39.44}	&37.41  
\\
    Meta-SR \cite{DBLP:conf/cvpr/HuM0WT019}   &41.29	&-	&38.12	&-   
    &37.47	&-	&33.99	&-  
    &35.79	&-	&32.32	&- 
    &35.85	&-	&32.98	&-
    &42.92	&-	&39.22	&-  
\\

     LIIF \cite{DBLP:conf/cvpr/ChenL021}            &41.22	&38.99	&38.08	&\underline{36.99}   
                                    &37.44	&35.31	&33.96	&\underline{32.95}  
                                    &35.75	&33.68	&32.28	&31.45 
                                    &36.70	&34.08	&32.84	&31.70
                                    &42.77	&40.19	&39.13	&\underline{37.69}  
                                \\
        SRWarp \cite{DBLP:conf/cvpr/SonL21}            &\underline{41.32}	&39.04	&38.02	&36.91   
                                        &\underline{37.52}	&35.41	&33.98	&32.91  
                                        &\underline{35.84}	&33.77	&32.31	&\underline{31.47} 
                                        &36.96	&\underline{34.46}	&33.04	&\underline{31.88}
                                        &\underline{43.16}	&40.30	&39.16	&\underline{37.69}  
                                        \\
        LTEW \cite{DBLP:conf/eccv/LeeCJ22}            &\textbf{41.52}	&\textbf{39.32}	&\underline{38.25}	&\textbf{37.14}   
                                        &\textbf{37.61}	&\textbf{35.54}	&\underline{34.01}	&\textbf{33.16}  
                                        &\textbf{35.91}	&\textbf{33.87}	&\underline{32.37}	&\textbf{31.55} 
                                        &\textbf{37.02}	&\textbf{34.63}	&\underline{33.11}	&\textbf{31.99}
                                        &\textbf{43.19}	&\textbf{40.73}	&\underline{39.41}	&\textbf{38.04}  
                        \\
 
                LeRF-Net++        &40.74	&38.78	&\textbf{38.27}	&36.47   
                &36.78	&35.05	&\textbf{34.12}	&32.59  
                &35.58	&33.47	&\textbf{32.41}	&31.25 
                &34.38	&33.56	&\textbf{33.35}	&30.87
                &41.52	&39.58	&\textbf{39.44}	&36.39  
                                        \\   
    \midrule \midrule
    \multirow{2}{*}{Method} &\multicolumn{4}{c}{Set5} &\multicolumn{4}{c}{Set14} &\multicolumn{4}{c}{BSDS100} &\multicolumn{4}{c}{Urban100} &\multicolumn{4}{c}{Manga109} \\ \cmidrule(lr){2-5} \cmidrule(lr){6-9} \cmidrule(lr){10-13} \cmidrule(lr){14-17} \cmidrule(lr){18-21} 
       &$\frac{\times2.0}{\times3.0}$   &$\frac{\times3.0}{\times3.0}$   &$\frac{\times3.0}{\times4.0}$   &$\frac{\times4.0}{\times4.0}$
    &$\frac{\times2.0}{\times3.0}$   &$\frac{\times3.0}{\times3.0}$   &$\frac{\times3.0}{\times4.0}$   &$\frac{\times4.0}{\times4.0}$
    &$\frac{\times2.0}{\times3.0}$   &$\frac{\times3.0}{\times3.0}$   &$\frac{\times3.0}{\times4.0}$   &$\frac{\times4.0}{\times4.0}$
    &$\frac{\times2.0}{\times3.0}$   &$\frac{\times3.0}{\times3.0}$   &$\frac{\times3.0}{\times4.0}$   &$\frac{\times4.0}{\times4.0}$
    &$\frac{\times2.0}{\times3.0}$   &$\frac{\times3.0}{\times3.0}$   &$\frac{\times3.0}{\times4.0}$   &$\frac{\times4.0}{\times4.0}$ \\ \midrule
                                     Nearest  &28.87	&27.91	&26.88	&26.25   
                                            &27.07	&26.08	&25.33	&24.74  
                                            &27.12	&26.17	&25.57	&25.03 
                                            &24.25	&23.34	&22.68	&22.17
                                            &26.12	&25.04	&24.06	&23.43  
                                    \\
                                   Bilinear   &30.43	&29.53	&28.27	&27.55
                                            &27.94	&27.04	&26.16	&25.51
                                            &27.60	&26.77	&26.11	&25.53
                                            &24.81	&23.99	&23.26	&22.68
                                            &27.21	&26.16	&24.95	&24.19
                                    \\
                                   Bicubic   &31.41	&30.39	&29.12	&28.42
                                            &28.70	&27.63	&26.75	&26.09
                                            &28.18	&27.20	&26.53	&25.95
                                            &25.43	&24.45	&23.71	&23.14
                                            &28.20	&26.95	&25.67	&24.90  
                                    \\
                                   Lanczos2  &31.44	&30.41	&29.14	&28.44
                                            &28.72	&27.64	&26.77	&26.10
                                            &28.20	&27.21	&26.54	&25.96
                                            &25.45	&24.47	&23.73	&23.15
                                            &28.23	&26.97	&25.70	&24.92
                                     \\ 
                                   Lanczos3  &31.85	&30.79	&29.49	&28.78
                                            &{29.04}	&27.91	&27.01	&26.31
                                            &28.43	&27.39	&26.70	&26.10
                                            &25.71	&24.68	&23.92	&23.32
                                            &28.70	&27.38	&26.02	&25.21
                                     \\
                                     RAISR*\cite{DBLP:journals/tci/RomanoIM17}       &32.35	&31.87	&29.91	&29.65   
                                     &29.04	&28.62	&27.10	&26.86  
                                     &28.24	&27.84	&26.68	&26.42 
                                     &25.92	&25.50	&24.08	&23.89
                                     &29.30	&28.73	&26.40	&26.12  
                               \\ 
                            LeRF-L       &31.95	&30.72	&29.73	&29.13   
                                    &29.15	&27.88	&27.13	&26.53  
                                    &28.63	&27.42	&26.84	&26.30 
                                    &25.86	&24.69	&24.05	&23.52
                                    &28.95	&27.43	&26.26	&25.50  
                                    \\
                                    LeRF-G       &\underline{33.17}	&\underline{32.02}	&\underline{30.86}	&\underline{30.15}   
                                    &\underline{30.06}	&\underline{28.84}	&\underline{28.05}	&\underline{27.35}  
                                    &\underline{29.15}	&\underline{28.00}	&\underline{27.31}	&\underline{26.70} 
                                    &\underline{26.90}	&\underline{25.68}	&\underline{24.88}	&\underline{24.23}
                                    &\underline{30.86}	&\underline{29.48}	&\underline{28.10}	&\underline{27.25}  
                                    \\ 
                                    LeRF-Net       &\textbf{34.21}	&\textbf{33.04}	&\textbf{31.95}	&\textbf{31.26}   
                                            &\textbf{30.79}	&\textbf{29.57}	&\textbf{28.79}	&\textbf{28.10}  
                                            &\textbf{29.76}	&\textbf{28.54}	&\textbf{27.84}	&\textbf{27.22} 
                                            &\textbf{28.14}	&\textbf{26.82}	&\textbf{25.91}	&\textbf{25.22}
                                            &\textbf{32.38}	&\textbf{30.97}	&\textbf{29.63}	&\textbf{28.79}  
                                            \\ 
                                    
                                    \midrule
RCAN* \cite{DBLP:conf/eccv/ZhangLLWZF18}                      &35.27	&34.08	&32.49	&31.67   
&31.62  	&30.19	&29.13	&28.36  
&30.35  	&29.01	&28.14	&27.43 
&29.67  	&27.81	&26.60	&25.70
&34.89  	&33.10	&30.77	&29.55  
\\
    Meta-SR \cite{DBLP:conf/cvpr/HuM0WT019}                      &-	&\underline{34.71}	&-	&32.48   
                                    &-  	&30.56	&-	&28.83  
                                    &-  	&29.26	&-	&27.73 
                                    &-  	&\underline{28.91}	&-	&26.69
                                    &-  	&\underline{34.37}	&-	&\textbf{31.32}  
                                \\
                                
    LIIF \cite{DBLP:conf/cvpr/ChenL021}            &\underline{35.63}	&34.59	&33.17	&32.37   
                                    &31.68	&30.39	&29.45	&28.65  
                                    &\underline{30.47}	&29.24	&28.42	&27.73 
                                    &30.23	&28.80	&27.55	&26.66
                                    &\underline{35.73}	&34.17	&\underline{32.29}	&\underline{31.19}  
                                \\ 
        SRWarp \cite{DBLP:conf/cvpr/SonL21}            &35.59	&34.54	&33.00	&32.05   
                                        &\underline{31.69}	&30.45	&29.46	&28.60  
                                        &30.46	&29.20	&28.33	&27.58 
                                        &\underline{30.37}	&\underline{28.91}	&27.60	&26.60
                                        &35.63	&34.14	&32.11	&30.73  
                                        \\ 

        LTEW \cite{DBLP:conf/eccv/LeeCJ22}            &\textbf{35.92}	&\textbf{34.75}	&\textbf{33.36}	&\textbf{32.54}   
                                        &\textbf{31.92}	&\textbf{30.60}	&\textbf{29.64}	&\textbf{28.89}  
                                        &\textbf{30.59}	&\textbf{29.31}	&\textbf{28.50}	&\textbf{27.78} 
                                        &\textbf{30.63}	&\textbf{29.05}	&\textbf{27.82}	&\textbf{26.88}
                                        &\textbf{36.17}	&\textbf{34.42}	&\textbf{32.54}	&\textbf{31.32}  
                                        \\
        LeRF-Net++            &35.33	&34.57	&\underline{33.27}	&\underline{32.52}   
                                        &31.57	&\underline{30.52}	&\underline{29.60}	&\underline{28.86}  
                                        &30.33	&\underline{29.27}	&\underline{28.45}	&\textbf{27.78} 
                                        &29.74	&28.82	&\underline{27.64}	&\underline{26.75}
                                        &34.61	&33.66	&32.06	&31.11
                                        \\                            
                                \bottomrule
    \end{tabular}%
    }

    \begin{tablenotes}
        \item $\frac{\times r_h}{\times r_w}$ denotes upsampling $r_h$ times along the short side and $r_w$ times along the long side. 
        \item * denotes that we combine fixed-scale super-resolution methods with Bicubic to achieve arbitrary-scale upsampling. 
        \item The best and second-best results are \textbf{highlighted} and \underline{underlined}. 
    \end{tablenotes}
\end{threeparttable}
    \label{tab:main}
    \end{table*}

%% file: figures/visual_upsample.tex
\newcommand{\name}{0}
\newcommand{\h}{0}
\newcommand{\w}{0.15}
\newcommand{\hrs}{0.36}
\newcommand{\vsp}{0 mm}

\newcommand{\lefth}{-3mm}
\newcommand{\hrh}{-2mm}

\newlength \g
%
\begin{figure*}[t]
	\scriptsize
	\centering
%
%

\renewcommand{\h}{0.105}
\renewcommand{\w}{0.105}
\renewcommand{\hrs}{0.234}

\renewcommand{\name}{visual/comparison/X1.50_2.00/Manga109-UltraEleven/Y700_X120/UltraEleven-}
\setlength{\g}{-4mm}
\begin{tabular}{cc}
	\hspace{\lefth}
	\begin{adjustbox}{valign=t}
		\begin{tabular}{c}
			\includegraphics[height=\hrs \textwidth, width=\hrs \textwidth]{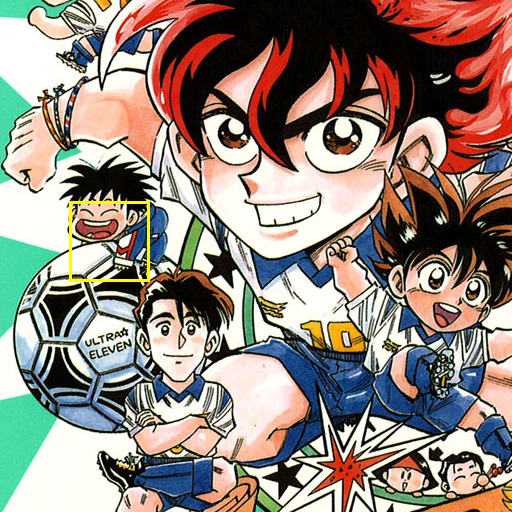}
			\\
			\scriptsize $\frac{\times1.5}{\times2.0}$  upsampling
		\end{tabular}
	\end{adjustbox}
	\hspace{\hrh}
	\begin{adjustbox}{valign=t}
		\begin{tabular}{ccccccc}
			\includegraphics[height=\h \textwidth, width=\w \textwidth]{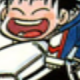}~\hspace{\g} &
			\includegraphics[height=\h \textwidth, width=\w \textwidth]{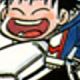}~\hspace{\g} &
			\includegraphics[height=\h \textwidth, width=\w \textwidth]{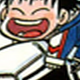}~\hspace{\g} & 
			\includegraphics[height=\h \textwidth, width=\w \textwidth]{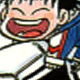}~\hspace{\g}  &
			\includegraphics[height=\h \textwidth, width=\w \textwidth]{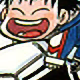}~\hspace{\g} & 
			\includegraphics[height=\h \textwidth, width=\w \textwidth]{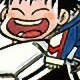}~
			\\
			Bilinear \hspace{\g} &
			Bicubic \hspace{\g} &
			Lanczos3 \hspace{\g} &
			LeRF-L \hspace{\g} &
			LeRF-G \hspace{\g} &
			LeRF-Net
			\\
			\includegraphics[height=\h \textwidth, width=\w \textwidth]{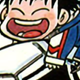}~\hspace{\g} &
			\includegraphics[height=\h \textwidth, width=\w \textwidth]{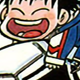}~\hspace{\g} & 
			\includegraphics[height=\h \textwidth, width=\w \textwidth]{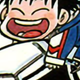}~\hspace{\g} &
			\includegraphics[height=\h \textwidth, width=\w \textwidth]{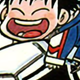}~\hspace{\g}  &
			\includegraphics[height=\h \textwidth, width=\w \textwidth]{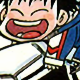}~\hspace{\g} & 
			\includegraphics[height=\h \textwidth, width=\w \textwidth]{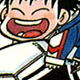}~
			\\
			RCAN* \hspace{\g} &
			LIIF \hspace{\g} &
			SRWarp \hspace{\g} &
			LTEW \hspace{\g} &
			LeRF-Net++ \hspace{\g} &
			GT
			\\
		\end{tabular}
	\end{adjustbox}
\end{tabular}

\vspace{\vsp}

\renewcommand{\name}{visual/comparison/X2.00_2.40/Manga109-MukoukizuNoChonbo/Y10_X710/MukoukizuNoChonbo-}
\setlength{\g}{-4mm}
\begin{tabular}{cc}
	\hspace{\lefth}
	\begin{adjustbox}{valign=t}
		\begin{tabular}{c}
			\includegraphics[height=\hrs \textwidth, width=\hrs \textwidth]{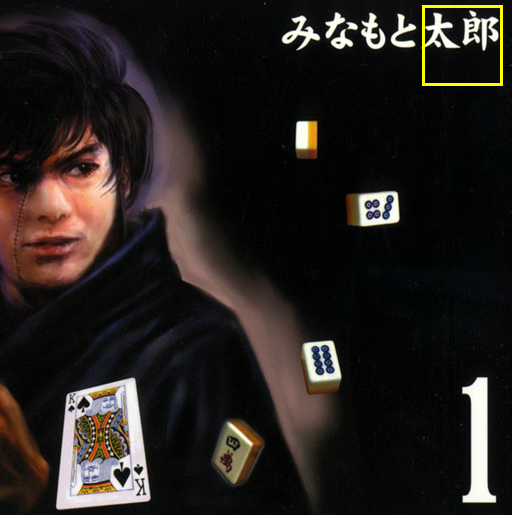}
			\\
			\scriptsize $\frac{\times2.0}{\times2.4}$ upsampling
		\end{tabular}
	\end{adjustbox}
	\hspace{\hrh}
	\begin{adjustbox}{valign=t}
		\begin{tabular}{ccccccc}
			\includegraphics[height=\h \textwidth, width=\w \textwidth]{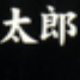}~\hspace{\g} &
			\includegraphics[height=\h \textwidth, width=\w \textwidth]{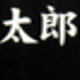}~\hspace{\g} &
			\includegraphics[height=\h \textwidth, width=\w \textwidth]{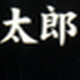}~\hspace{\g} & 
			\includegraphics[height=\h \textwidth, width=\w \textwidth]{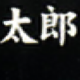}~\hspace{\g}  &
			\includegraphics[height=\h \textwidth, width=\w \textwidth]{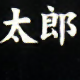}~\hspace{\g} & 
			\includegraphics[height=\h \textwidth, width=\w \textwidth]{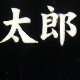}~
			\\
			Bilinear \hspace{\g} &
			Bicubic \hspace{\g} &
			Lanczos3 \hspace{\g} &
			LeRF-L \hspace{\g} &
			LeRF-G \hspace{\g} &
			LeRF-Net
			\\
			\includegraphics[height=\h \textwidth, width=\w \textwidth]{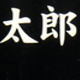}~\hspace{\g} &
			\includegraphics[height=\h \textwidth, width=\w \textwidth]{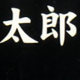}~\hspace{\g} & 
			\includegraphics[height=\h \textwidth, width=\w \textwidth]{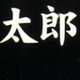}~\hspace{\g} &
			\includegraphics[height=\h \textwidth, width=\w \textwidth]{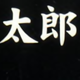}~\hspace{\g}  &
			\includegraphics[height=\h \textwidth, width=\w \textwidth]{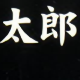}~\hspace{\g} & 
			\includegraphics[height=\h \textwidth, width=\w \textwidth]{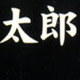}~
			\\
			RCAN* \hspace{\g} &
			LIIF \hspace{\g} &
			SRWarp \hspace{\g} &
			LTEW \hspace{\g} &
			LeRF-Net++ \hspace{\g} &
			GT
			\\
		\end{tabular}
	\end{adjustbox}
\end{tabular}

\vspace{\vsp}

\renewcommand{\name}{visual/comparison/X4.00_4.00/Urban100-img092/Y400_X260/img092-}
\setlength{\g}{-4mm}
\begin{tabular}{cc}
	\hspace{\lefth}
	\begin{adjustbox}{valign=t}
		\begin{tabular}{c}
			\includegraphics[height=\hrs \textwidth, width=\hrs \textwidth]{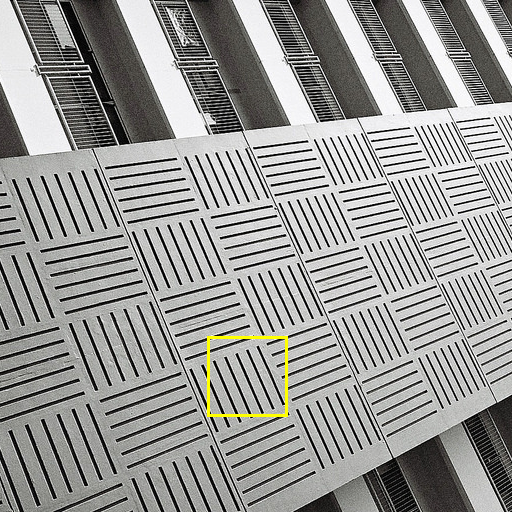}
			\\
			\scriptsize $\frac{\times4.0}{\times4.0}$ upsampling
		\end{tabular}
	\end{adjustbox}
	\hspace{\hrh}
	\begin{adjustbox}{valign=t}
		\begin{tabular}{ccccccc}
			\includegraphics[height=\h \textwidth, width=\w \textwidth]{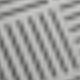}~\hspace{\g} &
			\includegraphics[height=\h \textwidth, width=\w \textwidth]{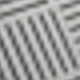}~\hspace{\g} &
			\includegraphics[height=\h \textwidth, width=\w \textwidth]{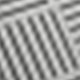}~\hspace{\g} & 
			\includegraphics[height=\h \textwidth, width=\w \textwidth]{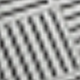}~\hspace{\g}  &
			\includegraphics[height=\h \textwidth, width=\w \textwidth]{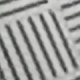}~\hspace{\g} & 
			\includegraphics[height=\h \textwidth, width=\w \textwidth]{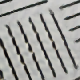}~
			\\
			Bilinear \hspace{\g} &
			Bicubic \hspace{\g} &
			Lanczos3 \hspace{\g} &
			LeRF-L \hspace{\g} &
			LeRF-G \hspace{\g} &
			LeRF-Net
			\\
			\includegraphics[height=\h \textwidth, width=\w \textwidth]{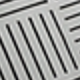}~\hspace{\g} &
			\includegraphics[height=\h \textwidth, width=\w \textwidth]{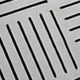}~\hspace{\g} & 
			\includegraphics[height=\h \textwidth, width=\w \textwidth]{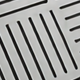}~\hspace{\g} &
			\includegraphics[height=\h \textwidth, width=\w \textwidth]{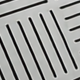}~\hspace{\g}  &
			\includegraphics[height=\h \textwidth, width=\w \textwidth]{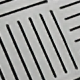}~\hspace{\g} & 
			\includegraphics[height=\h \textwidth, width=\w \textwidth]{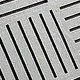}~
			\\
			RCAN* \hspace{\g} &
			LIIF \hspace{\g} &
			SRWarp \hspace{\g} &
			LTEW \hspace{\g} &
			LeRF-Net++ \hspace{\g} &
			GT
			\\
		\end{tabular}
	\end{adjustbox}
\end{tabular}

\vspace{\vsp}

\caption{Qualitative comparison for arbitrary-scale upsampling. From top to bottom, the example images are: \emph{UltraEleven} (Manga109), \emph{MukoukizuNoChonbo} (Manga109), and \emph{img092} (Ubran100). Best view in color and on screen.}
\label{fig:visual_upsample}
\end{figure*}

%% file: parts/experiments.tex
\section{Experiments and Results}

\subsection{Experimental Settings}

\noindent\textbf{Datasets and metrics.}
We train LeRF models on the DIV2K dataset \cite{DBLP:conf/cvpr/AgustssonT17}, which is widely used in image resampling tasks. The DIV2K dataset covers multiple scenes and encapsulates diverse image patches with various local structures. It contains 800 training images and 100 validation images at 2K resolution. We train LeRF models on the $\times4$ Bicubic-downsampled image pairs and apply the obtained models to arbitrary transformations, including two representative resampling tasks, \emph{i.e.}, arbitrary-scale upsampling and homographic warping. For arbitrary-scale upsampling, we evaluate LeRF with 6 benchmark datasets: Set5, Set14, BSDS100 \cite{DBLP:conf/iccv/MartinFTM01}, Urban100 \cite{DBLP:conf/cvpr/HuangSA15}, Manga109 \cite{DBLP:journals/mta/MatsuiIAFOYA17}, and DIV2K-Valid \cite{DBLP:conf/cvpr/AgustssonT17}. We select representative symmetric or asymmetric upsampling scales for evaluation and apply Bicubic interpolation as the degradation assumption to obtain the LR images. For homographic warping, we evaluate LeRF on DIV2K-Warping (DIV2KW) dataset proposed in SRWarp \cite{DBLP:conf/cvpr/SonL21}, which simulates a homographic transformation for each image in DIV2K-Valid. DIV2KW is further split into in-scale and out-scale parts by LTEW \cite{DBLP:conf/eccv/LeeCJ22} according to whether the transformation parameters are seen or not during the training process. As to the performance evaluation of arbitrary-scale upsampling, we report Y-Channel PSNR and SSIM \cite{DBLP:journals/tip/WangBSS04} for fidelity, and LPIPS \cite{DBLP:conf/cvpr/ZhangIESW18} for perceptual quality. For homographic warping, following SRWarp \cite{DBLP:conf/cvpr/SonL21} and LTEW \cite{DBLP:conf/eccv/LeeCJ22}, we report masked PSNR (mPSNR), which is measured by averaging PSNR across three color channels in valid regions. As to efficiency, we report running time (mobile CPU, desktop CPU, and desktop GPU), theoretical multiply-accumulate operations (MACs), and storage requirements to evaluate the performance-efficiency trade-off.

\noindent\textbf{Comparison methods.}
The main competitors to our efficiency-orientated models are interpolation methods, including Nearest, Bilinear, Bicubic, Lanczos2, and Lanczos3. Furthermore, we combine fixed-scale super-resolution methods with Bicubic interpolation as additional baselines for arbitrary-scale upsampling (RAISR*). Specifically, since adaptive filtering methods (Fig.~\ref{fig:overview}(b)) such as RAISR \cite{DBLP:journals/tci/RomanoIM17} lack generalization ability across different upsampling scales, we fuse the results of $\times 2$, $\times 3$, and $\times 4$ models by choosing the model with the closest integer upsampling scales (\emph{e.g.}, $\times 3$ model for  $\frac{\times2.0}{\times3.0}$). To evaluate our performance-orientated model, we include representative DNN-based arbitrary-scale upsampling methods (Meta-SR \cite{DBLP:conf/cvpr/HuM0WT019} and LIIF \cite{DBLP:conf/cvpr/ChenL021}) and homographic warping methods (SRWarp \cite{DBLP:conf/cvpr/SonL21} and LTEW \cite{DBLP:conf/eccv/LeeCJ22}). We also include a fixed-scale super-resolution model by cascading the pre-trained $\times 2$ upsampler with Bicubic interpolation (RCAN*).

\noindent\textbf{Implementation details.}
Our method is trained with the Adam optimizer \cite{DBLP:journals/corr/KingmaB14} in the cosine annealing schedule \cite{DBLP:conf/iclr/LoshchilovH17}. We train LeRF with the MSE loss function for $5 \times 10^4$ iterations at a batch size of 32. For both amplified linear and anisotropic Gaussian, we choose a $2 \times 2$ support patch ($\Omega_p$ in Eq.~\ref{eq:lerf}). To obtain a fair comparison in running time tests, we implement LeRF and interpolation methods on the Android platform under the same JAVA $\mathtt{IntStream.parallel()}$ API, with the only difference being the resampling functions and LUT retrieval or network inference in our method. We estimate the MACs of LeRF and interpolation methods based on the size of the target image and the operations needed per target pixel. Specifically, for the $\mathtt{exp()}$ operation in LeRF and $\mathtt{sinc()}$ operation in Lanczos interpolation, we count their MAC as 4, according to the theoretical estimation in \cite{10.2307/2030559} and the decimal precision of $\mathtt{float}$ type number. For DNN-based methods, we calculate their MACs with $\mathtt{torchinfo}$\footnote{\url{https://github.com/TylerYep/torchinfo}}. Our code is available in both MindSpore\footnote{\url{https://gitee.com/mindspore/models/tree/master/research/cv/lerf}} and Pytorch\footnote{\url{https://github.com/ddlee-cn/LeRF-PyTorch}}.

\input{tables/supp_main_table.tex}

\input{tables/comp_table.tex}


\subsection{Evaluation for Arbitrary-scale Upsampling}

\noindent\textbf{Quantitative comparison.} 
We list the quantitative comparisons for arbitrary-scale upsampling in TABLE~\ref{tab:main}, TABLE~\ref{tab:main-supp}, TABLE~\ref{tab:comp}, and TABLE~\ref{tab:lpips}. As can be seen, efficiency-orientated LeRF models (LeRF-L, LeRF-G, and LeRF-Net) achieve significantly better performance than interpolation methods. For example, LeRF-G exceeds Bicubic interpolation up to 3dB PSNR when upsampling $\frac{\times2.0}{\times2.0}$ on the Manga109 dataset. Performance-orientated LeRF model (LeRF-Net++) achieves comparable performance with existing DNNs.
Notably, efficiency-orientated LeRF models outperform interpolation methods in LPIPS by a large margin and achieve comparable (sometimes even better) performance with DNN-based methods, showing its clear advantage in perceptual quality.

\input{tables/lpips_table.tex}

\setcounter{figure}{8}
\input{figures/fig_continuous.tex}
\setcounter{figure}{7}
\input{figures/visual_rule.tex}
\setcounter{figure}{9}

\noindent\textbf{Qualitative comparison.} 
In Fig.~\ref{fig:visual_upsample}, we compare the visual quality of LeRF models with interpolation and DNN-based methods. As can be seen, our efficiency-orientated models obtain better visual quality over interpolation across various local structures and textures. Specifically, since the resampling functions are adapted to local structures, LeRF models are capable of retaining clearer textures and avoiding blurry boundaries. LeRF-Net++ obtains similar visual quality compared with representative DNNs. In Fig.~\ref{fig:visual_rule}, we compare LeRF-G with two rule-based adaptive interpolation methods, \emph{i.e.}, NEDI \cite{DBLP:journals/tip/LiO01a} and SKR \cite{DBLP:journals/tip/TakedaFM07}. They estimate the resampling weights according to hand-designed rules based on local gradients. As can be seen, LeRF-G obtains better visual quality than these rule-based adaptive resampling methods, showing the advantage of extracting structural priors in a learning-based way.

\noindent\textbf{Continuous Resampling Results.}
In Fig.~\ref{fig:continuous}, we provide an example of continuous upsampling results of LeRF-G. Besides, on our project page\footnote{\url{https://lerf.pages.dev}}, we present two comparison videos of continuous resampling results. In the first one, we compare the visual results of LeRF-G and Bicubic side-by-side for a continuous upsampling range, \emph{i.e.}, from $\times1$ to $\times8$. In the second one, we compare the continuous resampling results of LeRF-G and Bicubic for homographic warping transformations, including asymmetric upsampling, downsampling, sheering, and rotation. As can be seen, LeRF-G produces more visually pleasing results than the widely used Bicubic interpolation.

\input{figures/visual_warp.tex}

\subsection{Evaluation for Homographic Warping}

\noindent\textbf{Quantitative comparison.} 
We list the quantitative comparisons for homographic warping in TABLE~\ref{tab:comp}. As can be seen, our efficiency-orientated models (LeRF-L, LeRF-G, and LeRF-Net) achieve better performance in both in-scale and out-scale transformations over interpolation methods. For example, LeRF-G exceeds Bicubic interpolation over 1dB mPSNR on the in-scale split of DIV2KW dataset. Our performance-orientated model (LeRF-Net++) achieves comparable performance with tailored homographic warping DNNs at much higher efficiency (\emph{e.g.}, 130ms vs. 757ms of LTEW).

\noindent\textbf{Qualitative comparison.} 
In Fig.~\ref{fig:visual_warp}, we show the visual quality of the warping results of LeRF models and compare them with those of Bicubic interpolation. As can be seen, LeRF-G obtains clearly better visual quality than Bicubic interpolation. LeRF-Net and LeRF-Net++ further improve the sharpness of edges and reduce aliasing artifacts.

\input{figures/fig_tradeoff.tex}

\subsection{Efficiency Evaluation}
To evaluate the efficiency, we conduct running time tests, estimate the theoretical MACs, and report storage requirements for LeRF models and comparison methods. We list the detailed comparison in TABLE~\ref{tab:comp}. As can be seen, LeRF-L and LeRF-G achieve comparable or better running time compared to the popular interpolation methods on both mobile CPU and desktop GPU, showing the potential that LeRF could play the role as a superior competitor with significantly better performance. Compared to DNN-based methods, our efficiency-orientated models, \emph{i.e.}, LeRF-L, LeRF-G, and LeRF-Net show a clear advantage in efficiency on desktop GPU (\emph{e.g.}, 15ms of LeRF-Net vs. 677ms of LIIF). On the other hand, LeRF-Net++ achieves similar performance compared to DNNs at higher efficiency and less extra storage (\emph{e.g.}, 65.99MB vs. 196.13MB of LTEW), demonstrating its practicality for being deployed on various devices. Finally, the advantage of LeRF in terms of performance-efficiency trade-off is also validated by the results in trade-off planes in Fig.~\ref{fig:tradeoff_cpu} (CPU running time \& upsampling PSNR) and Fig.~\ref{fig:tradeoff} (GPU running time \& warping mPSNR).

\input{figures/visual_affine.tex}
\input{figures/visual_down.tex}

\subsection{Generalization Evaluation}

In Fig.~\ref{fig:visual_affine}, we evaluate the generalization ability of LeRF to arbitrary transformations and compare the results of LeRF-G with widely applied interpolation methods. As can be seen, the learned resampling functions generalize well to unseen deformations, \emph{e.g.}, barrel-shape warping and warping according to optical flow. LeRF generates sharper edges and retains more texture details, leading to more visually pleasing results than popular interpolation methods. In our previous work \cite{lerf_cvpr}, we show a limitation of LeRF that aliasing artifacts appear when downsampling highly dense textures. This phenomenon can be significantly alleviated by replacing the pre-processing network with an Anti-Aliasing (AA) filter, \emph{i.e.}, an isotropic Gaussian filter. In Fig.~\ref{fig:visual_down}, we show the effectiveness of this AA filter, demonstrating the flexibility of LeRF by introducing a configurable pre-processing stage.

\subsection{Ablation Analysis}

\input{figures/fig_hyper_vis.tex}

\input{figures/fig_optim_kernel.tex}

\noindent\textbf{Visualization of learned resampling functions.} 
We further visualize the intermediate results of LeRF-G in Fig.~\ref{fig:hyper_vis}. As can be seen, the hyper-parameter $\rho$ clearly distinguishes the orientations of edges (red vs. blue), and $\sigma_X$ and $\sigma_Y$ capture the horizontal and vertical lines, respectively. In addition, in the second row of each example, we visualize the resampling functions defined by the predicted hyper-parameters. The shapes of resampling functions are well adapted to corners, flat surfaces, and edges with various orientations. These results validate the effectiveness of extracting structural priors in a data-driven way.

\noindent\textbf{Analysis of GT-optimized resampling functions.}
To analyze the optimal shapes of resampling functions across different transformations, we obtain hyper-parameters in resampling functions via a per-image optimization process. In Fig.~\ref{fig:optim_kernel}, we visualize the clustering results on Set14 of these GT-optimized resampling functions across different upsampling scales. As can be seen, the GT-optimized cluster centroids are very similar across different upsampling scales, explaining the generalization ability of our method.

\input{tables/abl_table1.tex}

\input{figures/visual_abl1.tex}

\noindent\textbf{The effectiveness of the adaptive resampling function.}
We conduct the following ablation experiments on the Set5 benchmark to analyze the design of the adaptive resampling function. 1) Fixed Gaussian: We freeze the hyper-parameters to $\rho=0,\sigma_X = \sigma_Y = 1$ in the anisotropic Gaussian. 2) Isotropic Keys $\mathtt{cubic}$: We utilize the isotropic Keys $\mathtt{cubic}$ function \cite{Keys1981CubicCI} (Eq.~\ref{eq:cubic}) as an adaptive resampling function, which has only one hyper-parameter $a$ that controls the sharpness of the one-dimension piecewise $\mathtt{cubic}$. The experimental results are listed in TABLE~\ref{tab:abl1}. As can be seen, LeRF-L could achieve similar performance with isotropic Keys $\mathtt{cubic}$ with fewer MACs. Besides, we show the visual results in Fig.~\ref{fig:visual_abl1}, non-steerable resampling functions like isotropic Keys $\mathtt{cubic}$ produce zig-zag artifacts, showing the effectiveness of the anisotropic resampling functions.

\input{tables/abl_table3.tex}
\input{figures/visual_abl2.tex}

\noindent\textbf{The effectiveness of the LUT acceleration and the proposed adaptations.}
As listed in TABLE~\ref{tab:abl3}, we conduct the following ablation experiments to analyze the design of the LUT acceleration and our adaptations. 1) Without LUT acceleration: Compared with the multi-branch DNN (Fig.~\ref{fig:net_arch}), the LUT acceleration saves a lot of computations at a very small performance cost. 2) Without DE: The visualization of the learned hyper-parameter $\rho$ in Fig.~\ref{fig:visual_abl2} shows that RE makes an isotropic assumption and thus lacks the ability to learn edge orientations, resulting in a performance drop. 3) Without edge-sensitive patterns: LeRF-G with ``SCX'' indexing patterns outperforms the variant with only the default ``S'' pattern, showing the ability of edge-sensitive indexing patterns to better capture edge structures.

\input{tables/abl_net_table.tex}

\input{figures/visual_abl_pre.tex}

\noindent\textbf{Analysis of the pre-processing stage.}
We visualize the intermediate results of different pre-processing stages in Fig.~\ref{fig:visual_abl_pre}. As can be seen, the pre-processing network in LeRF-L, LeRF-G, and LeRF-Net enhances the edges of the input image for the resampling step. As listed in TABLE~\ref{tab:abl3}, this enhancement improves the resampling performance. LeRF-Net++ takes advantage of a pre-trained upsampling backbone and pre-upsample the image, resulting in clearer details for cascaded resampling. LeRF-G with AA filter blurs the image to reduce aliasing artifacts when downsampling.

\noindent\textbf{Analysis of network architecture.}
In TABLE~\ref{tab:abl_net}, we further investigate the network architecture of DNN used in LeRF-G and LeRF-Net. To utilize LUT acceleration, we keep the branches in the multi-branch DNN of LeRF-G as the same architecture of the network in previous LUT-based fixed-scale super-resolution methods like SR-LUT \cite{DBLP:conf/cvpr/JoK21} and MuLUT \cite{mulut, mulut23} (see Fig.~\ref{fig:net_arch}), which has a restricted receptive field (\emph{i.e.}, the size of $\mathcal{N}_p$ in Eq.~\ref{eq:training}) to avoid a substantial increase in LUT size. In LeRF-Net, we release this constraint in network architecture introduced by LUT acceleration and present an IMDN-like lightweight DNN for hyper-parameter learning and pre-processing. As listed in TABLE~\ref{tab:abl_net}, with six convolution (conv. for short) layers with a $3\times3$ kernel size, a simple DNN with $13\times13$ RF is constructed, achieving better performance than LeRF-G. With IMDN-like lightweight weight DNN in LeRF-Net, the receptive field can be further enlarged to  $43\times43$. As listed in TABLE~\ref{tab:abl_net}, LeRF-Net benefits from larger RF and increased parameter capacity. But, when the channel number (nf. for short) exceeds 64, the performance gain is marginal. We attribute this phenomenon to the strong assumption of explicit resampling function, and thus we shift to improve the pre-processing stage by proposing LeRF-Net++ for further scaling our LeRF method.

\input{tables/abl_table2.tex}

\input{tables/comparison_lut.tex}

\noindent\textbf{Analysis of training data.}
By default, LeRF is trained with $\times 4$ downsampled data pairs. We retrain our method on $\times2$ data pairs and mixed-scale data pairs with both symmetric and asymmetric upsampling scales. As listed in TABLE~\ref{tab:abl2}, training on mixed-scale data pairs yields similar performance, showing the robustness and generalization ability of our method, which is consistent with the clustering results in Fig.~\ref{fig:optim_kernel}.

\noindent\textbf{Additional comparison with LUT-based super-resolution methods.} 
We conduct an experiment to compare LeRF with LUT-based super-resolution methods, and the results are listed in TABLE~\ref{tab:comparison_lut}. As can be seen, retraining SR-LUT with Bicubic on mixed-scale LR-HR data pairs yields inferior performance. When incorporating LUT-based super-resolution methods as the pre-processing stage, LeRF-G achieves similar or better performance at a comparable efficiency with Bicubic interpolation. Thus, LeRF shows promising potential to work with not only performance-orientated upsamplers like RCAN \cite{DBLP:conf/cvpr/ZhangTKZ018}, but also efficient-orientated upsamplers such as SR-LUT \cite{DBLP:conf/cvpr/JoK21} and MuLUT \cite{mulut, mulut23}.

%% file: tables/supp_main_table.tex

\definecolor{Gray}{gray}{.95}
\renewcommand{\arraystretch}{1.2}

\begin{table*}[htbp]
        \caption{Quantitative comparison in SSIM for arbitrary-scale upsampling.}
    \begin{threeparttable}
        \resizebox{\textwidth}{!}{%
    \begin{tabular}{lcccccccccccccccccccc}
    \toprule
    \multirow{2}{*}{Method} &\multicolumn{4}{c}{Set5} &\multicolumn{4}{c}{Set14} &\multicolumn{4}{c}{BSDS100} &\multicolumn{4}{c}{Urban100} &\multicolumn{4}{c}{Manga109} \\ \cmidrule(lr){2-5} \cmidrule(lr){6-9} \cmidrule(lr){10-13} \cmidrule(lr){14-17} \cmidrule(lr){18-21} 
                           &$\frac{\times1.5}{\times1.5}$   &$\frac{\times1.5}{\times2.0}$   &$\frac{\times2.0}{\times2.0}$   &$\frac{\times2.0}{\times2.4}$
                        &$\frac{\times1.5}{\times1.5}$   &$\frac{\times1.5}{\times2.0}$   &$\frac{\times2.0}{\times2.0}$   &$\frac{\times2.0}{\times2.4}$
                        &$\frac{\times1.5}{\times1.5}$   &$\frac{\times1.5}{\times2.0}$   &$\frac{\times2.0}{\times2.0}$   &$\frac{\times2.0}{\times2.4}$
                        &$\frac{\times1.5}{\times1.5}$   &$\frac{\times1.5}{\times2.0}$   &$\frac{\times2.0}{\times2.0}$   &$\frac{\times2.0}{\times2.4}$
                        &$\frac{\times1.5}{\times1.5}$   &$\frac{\times1.5}{\times2.0}$   &$\frac{\times2.0}{\times2.0}$   &$\frac{\times2.0}{\times2.4}$ \\ \midrule
     Nearest  &.9188	&.9086	&.9001	&.8747   
                                            &.8774	&.8631	&.8466	&.8168  
                                            &.8626	&.8456	&.8255	&.7944 
                                            &.8595	&.8405	&.8213	&.7871
                                            &.9281	&.9187	&.9103	&.8829  
                                    \\
                                    Bilinear   &.9491	&.9287	&.9120	&.8991
                                            &.9052	&.8745	&.8411	&.8240
                                            &.8868	&.8497	&.8110	&.7919
                                            &.8844	&.8451	&.8094	&.7886
                                            &.9573	&.9327	&.9132	&.8963
                                    \\
                                    Bicubic    &.9621	&.9446	&.9306	&.9174
                                            &.9277	&.9004	&.8703	&.8515
                                            &.9142	&.8802	&.8440	&.8218
                                            &.9102	&.8741	&.8410	&.8173
                                            &.9718	&.9514	&.9353	&.9175   \\ 
                                Lanczos2  &.9624	&.9450	&.9309	&.9176
                                        &.9283	&.9012	&.8712	&.8522
                                        &.9150	&.8811	&.8451	&.8227
                                        &.9110	&.8750	&.8419	&.8180
                                        &.9722	&.9519	&.9359	&.9179
                                \\ 
                                    Lanczos3  &.9666	&.9499	&.9366	&.9234
                                            &.9359	&.9096	&.8804	&.8614
                                            &.9245	&.8911	&.8554	&.8327
                                            &.9194	&.8836	&.8509	&.8268
                                            &.9772	&.9578	&.9428	&.9246
                                     \\ 
                                RAISR* \cite{DBLP:journals/tci/RomanoIM17}      &.9516	&.9496	&\underline{.9481}	&.9266   
                                &.9121	&.9051	&.8990	&.8701  
                                &.8987	&.8875	&.8787	&.8448 
                                &.8956	&.8872	&.8831	&.8423
                                &.9621	&.9599	&\underline{.9593}	&.9299  
                                \\ 
                                LeRF-L       &.9698	&.9551	&.9432	&.9301   
                                &.9447	&.9205	&.8937	&.8755  
                                &.9374	&.9068	&.8729	&.8517 
                                &.9313	&.8976	&.8660	&.8428
                                &.9794	&.9632	&.9501	&.9332  
                                \\
                                LeRF-G       &\underline{.9702}	&\underline{.9574}	&.9474	&\underline{.9376}   
                                &\underline{.9467}	&\underline{.9243}	&\underline{.8999}	&\underline{.8830}  
                                &\underline{.9391}	&\underline{.9104}	&\underline{.8789}	&\underline{.8580} 
                                &\underline{.9408}	&\underline{.9117}	&\underline{.8846}	&\underline{.8626}
                                &\underline{.9800}	&\underline{.9676}	&.9580	&\underline{.9462}  
                                \\ 
                                       LeRF-Net       &\textbf{.9745}	&\textbf{.9641}	&\textbf{.9562}	&\textbf{.9478}   
                                              &\textbf{.9535}	&\textbf{.9331}	&\textbf{.9111}	&\textbf{.8959}  
                                              &\textbf{.9474}	&\textbf{.9214}	&\textbf{.8926}	&\textbf{.8726} 
                                              &\textbf{.9541}	&\textbf{.9322}	&\textbf{.9118}	&\textbf{.8929}
                                              &\textbf{.9854}	&\textbf{.9769}	&\textbf{.9703}	&\textbf{.9616}  
                                              \\ 
                                       
                                       \midrule
                                RCAN* \cite{DBLP:conf/eccv/ZhangLLWZF18}                      &.9779	&.9687	&\textbf{.9616}	&.9535   
                                &.9591  	&.9419	&\textbf{.9218}	&.9067  
                                &.9527  	&.9290	&\textbf{.9029}	&.8823 
                                &.9673  	&.9525	&\textbf{.9386}	&.9182
                                &.9900  	&\underline{.9838}	&\textbf{.9788}	&.9715  
                            \\
                                MetaSR \cite{DBLP:conf/cvpr/HuM0WT019}                       &\underline{.9785}	&-	&.9610	&-   
                                        &\underline{.9602}	&-	&.9204	&-  
                                        &.9544	&-	&.9009	&- 
                                        &.9653	&-	&.9360	&-
                                        &.9904	&-	&.9872	&-  
                                        \\
                                        LIIF \cite{DBLP:conf/cvpr/ChenL021}            &.9784	&\underline{.9685}	&.9611	&\underline{.9538}   
                                        &.9600	&.9416	&\underline{.9210}	&\underline{.9074}  
                                        &.9545	&.9290	&.9011	&.8823
                                        &.9692	&.9505	&.9352	&.9206
                                        &.9903	&.9835	&.9781	&\underline{.9718}  
                                \\
                                SRWarp \cite{DBLP:conf/cvpr/SonL21}               &\underline{.9785}	&\underline{.9685}	&.9610	&.9533   
                                                        &\textbf{.9606}	&\underline{.9425}	&\underline{.9210}	&.9064  
                                                        &\underline{.9549}	&\underline{.9294}	&.9018	&\underline{.8829} 
                                                        &\underline{.9701}	&\underline{.9528}	&\underline{.9374}	&\underline{.9230}
                                                        &\underline{.9906}	&.9834	&.9780	&.9717  
                                                        \\
                        
                                LTEW \cite{DBLP:conf/eccv/LeeCJ22}            &\textbf{.9787}	&\textbf{.9690}	&\underline{.9615}	&\textbf{.9547}   
                                                                &\textbf{.9606}	&\textbf{.9427}	&\textbf{.9218}	&\textbf{.9089}  
                                                                &\textbf{.9550}	&\textbf{.9301}	&\underline{.9019}	&\textbf{.8833} 
                                                                &\textbf{.9702}	&\textbf{.9534}	&.9372	&\textbf{.9232}
                                                                &\textbf{.9907}	&\textbf{.9841}	&\underline{.9787}	&\textbf{.9728}  
                                                                \\
                                LeRF-Net++       &.9777	&.9669	&\textbf{.9616}	&.9521   
                                &.9597	&.9380	&\textbf{.9218}	&.9058  
                                &.9543	&.9237	&\textbf{.9029}	&.8822 
                                &.9632	&.9475	&\textbf{.9386}	&.9165
                                &.9895	&.9823	&\textbf{.9788}	&.9689  
                                \\
    \midrule \midrule
    \multirow{2}{*}{Method} &\multicolumn{4}{c}{Set5} &\multicolumn{4}{c}{Set14} &\multicolumn{4}{c}{BSDS100} &\multicolumn{4}{c}{Urban100} &\multicolumn{4}{c}{Manga109} \\ \cmidrule(lr){2-5} \cmidrule(lr){6-9} \cmidrule(lr){10-13} \cmidrule(lr){14-17} \cmidrule(lr){18-21} 
       &$\frac{\times2.0}{\times3.0}$   &$\frac{\times3.0}{\times3.0}$   &$\frac{\times3.0}{\times4.0}$   &$\frac{\times4.0}{\times4.0}$
    &$\frac{\times2.0}{\times3.0}$   &$\frac{\times3.0}{\times3.0}$   &$\frac{\times3.0}{\times4.0}$   &$\frac{\times4.0}{\times4.0}$
    &$\frac{\times2.0}{\times3.0}$   &$\frac{\times3.0}{\times3.0}$   &$\frac{\times3.0}{\times4.0}$   &$\frac{\times4.0}{\times4.0}$
    &$\frac{\times2.0}{\times3.0}$   &$\frac{\times3.0}{\times3.0}$   &$\frac{\times3.0}{\times4.0}$   &$\frac{\times4.0}{\times4.0}$
    &$\frac{\times2.0}{\times3.0}$   &$\frac{\times3.0}{\times3.0}$   &$\frac{\times3.0}{\times4.0}$   &$\frac{\times4.0}{\times4.0}$ \\ \midrule
     Nearest  &.8507	&.8128	&.7701	&.7372   
                                            &.7904	&.7355	&.6952	&.6553  
                                            &.7665	&.7082	&.6701	&.6307 
                                            &.7564	&.7006	&.6562	&.6166
                                            &.8572	&.8172	&.7748	&.7420  
                                    \\
                                    Bilinear   &.8774	&.8512	&.8155	&.7885
                                            &.7987	&.7556	&.7185	&.6824
                                            &.7652	&.7197	&.6837	&.6479
                                            &.7591	&.7152	&.6730	&.6363
                                            &.8706	&.8390	&.7984	&.7677
                                    \\
                                    Bicubic   &.8951	&.8686	&.8355	&.8106
                                            &.8241	&.7765	&.7410	&.7056
                                            &.7916	&.7399	&.7049	&.6694
                                            &.7848	&.7359	&.6954	&.6592
                                            &.8902	&.8572	&.8183	&.7888   \\ 
                                Lanczos2  &.8952	&.8687	&.8355	&.8107
                                        &.8247	&.7768	&.7412	&.7058
                                        &.7923	&.7402	&.7053	&.6698
                                        &.7854	&.7362	&.6956	&.6594
                                        &.8905	&.8574	&.8184	&.7889
                                \\
                                    Lanczos3  &.9010	&.8750	&.8413	&.8168
                                            &.8338	&.7855	&.7493	&.7130
                                            &.8018	&.7488	&.7130	&.6763
                                            &.7940	&.7443	&.7030	&.6659
                                            &.8967	&.8637	&.8238	&.7939
                                     \\
                                RAISR* \cite{DBLP:journals/tci/RomanoIM17}    &.9074	&.8968	&.8536	&.8431   
                                &.8311	&.8087	&.7584	&.7357 
                                &.8000	&.7729	&.7212	&.6963 
                                &.8013	&.7796	&.7179	&.6983
                                &.9039	&.8918	&.8370	&.8240  
                           \\ 
                                           LeRF-L       &.9059	&.8773	&.8491	&.8270   
                                           &.8476	&.7995	&.7653	&.7290  
                                           &.8211	&.7686	&.7327	&.6950 
                                           &.8095	&.7594	&.7197	&.6827
                                           &.9047	&.8709	&.8349	&.8069  
                                           \\
                                           LeRF-G       &\underline{.9189}	&\underline{.8980}	&\underline{.8732}	&\underline{.8548}   
                                           &\underline{.8573}	&\underline{.8126}	&\underline{.7816}	&\underline{.7475}  
                                           &\underline{.8278}	&\underline{.7763}	&\underline{.7412}	&\underline{.7047} 
                                           &\underline{.8309}	&\underline{.7844}	&\underline{.7462}	&\underline{.7114}
                                           &\underline{.9246}	&\underline{.9008}	&\underline{.8708}	&\underline{.8482}  
                                           \\ 
                                              LeRF-Net       &\textbf{.9332}	&\textbf{.9162}	&\textbf{.8964}	&\textbf{.8810}   
                                                 &\textbf{.8724}	&\textbf{.8307}	&\textbf{.8020}	&\textbf{.7702}  
                                                 &\textbf{.8438}	&\textbf{.7940}	&\textbf{.7601}	&\textbf{.7248} 
                                                 &\textbf{.8649}	&\textbf{.8240}	&\textbf{.7882}	&\textbf{.7558}
                                                 &\textbf{.9456}	&\textbf{.9275}	&\textbf{.9041}	&\textbf{.8861}  
                                                 \\ 
                                              
                                              \midrule
                                              RCAN* \cite{DBLP:conf/eccv/ZhangLLWZF18}                      &.9404	&.9252	&.9034	&.8871   
                                              &.8848  	&.8430	&.8121	&.7798  
                                              &.8547  	&.8060	&.7712	&.7355 
                                              &.8918  	&.8502	&.8131	&.7790
                                              &.9585  	&.9429	&.9177	&.8984
                                   \\
     MetaSR \cite{DBLP:conf/cvpr/HuM0WT019}                &-	&\underline{.9295}	&-	&\underline{.8985}   
                                     &-  	&.8466	&-	&.7876  
                                     &-  	&.8089	&-	&.7416 
                                     &-  	&.8672	&-	&.8049
                                     &-  	&\underline{.9491}	&-	&\underline{.9175}  
                                 \\
     LIIF \cite{DBLP:conf/cvpr/ChenL021}            &\underline{.9422}	&.9291	&\underline{.9114}	&.8981   
                                     &.8859	&\underline{.8472}	&\underline{.8189}	&.7878  
                                     &.8558	&.8101	&.7769	&.7425 
                                     &.8979	&.8664	&.8335	&.8043
                                     &\underline{.9612}	&.9487	&.9310	&.9169  
                                \\ 
                                SRWarp \cite{DBLP:conf/cvpr/SonL21}     &.9417	&.9286	&.9095	&.8953   
                                &.8842	&.8460	&.8181	&.7883  
                                &\underline{.8559}	&.8094	&.7767	&.7437 
                                &\underline{.8999}	&\underline{.8687}	&\underline{.8363}	&\underline{.8075}
                                &.9609	&.9485	&.9304	&.9159   
                                \\
                        
                LTEW \cite{DBLP:conf/eccv/LeeCJ22}            &\textbf{.9435}	&\textbf{.9301}	&\textbf{.9123}	&\textbf{.8989}   
                                                                &\underline{.8875}	&\textbf{.8481}	&\textbf{.8201}	&\textbf{.7894}  
                                                                &\textbf{.8573}	&\textbf{.8114}	&\textbf{.7786}	&.7438 
                                                                &\textbf{.9022}	&\textbf{.8698}	&\textbf{.8383}	&\textbf{.8093}
                                                                &\textbf{.9629}	&\textbf{.9500}	&\underline{.9329}	&\textbf{.9186}   
                                                                \\
                LeRF-Net++            &.9410	&.9289	&\underline{.9114}	&.8983   
                                                                &\textbf{.8885}	&.8467	&.8187	&\underline{.7886}  
                                                                &\underline{.8559}	&\underline{.8104}	&\underline{.7778}	&\textbf{.7440} 
                                                                &.8961	&.8672	&.8352	&.8057
                                                                &.9592	&.9476	&\textbf{.9393}	&.9167   
                                        \\\bottomrule
    \end{tabular}%
    }
    \begin{tablenotes}
        \item $\frac{\times r_h}{\times r_w}$ denotes upsampling $r_h$ times along the short side and $r_w$ times along the long side. 
        \item * denotes that we combine fixed-scale super-resolution methods with Bicubic to achieve arbitrary-scale upsampling. 
        \item The best and second-best results are \textbf{highlighted} and \underline{underlined}. 
    \end{tablenotes}
\end{threeparttable}
    \label{tab:main-supp}
    \end{table*}

%% file: tables/comp_table.tex

\renewcommand{\arraystretch}{1}
\begin{table*}[htbp]
    \centering
    \caption{Efficiency comparison of running time, MACs, and storage requirements, upsampling performance comparison on DIV2K-Valid dataset, and homographic warping performance comparison on DIV2KW dataset (In-scale and Out-scale).}
    \resizebox{0.8\textwidth}{!}{%
    \begin{threeparttable}
    \begin{tabular}{lrrrr|rrrrr}
        \toprule
          Method & \specialcell[r]{RunTime \\(CPU, ms)}  
          & \specialcell[r]{RunTime \\(GPU, ms)} 
          & \specialcell[r]{MACs}         
                & \specialcell[r]{Storage \\ Size}    
                & \specialcell[|r]{$\times 2$ Up \\ cPSNR}
                & \specialcell[r]{$\times 3$ Up \\ cPSNR} 
                & \specialcell[r]{$\times 4$ Up \\ cPSNR}
                & \specialcell[r]{In-scale \\ mPSNR}  
                & \specialcell[r]{Out-scale \\ mPSNR} \\
    \midrule
    Nearest                  &11  &3 &-       &-      &29.24    &26.66    &25.28 &25.95 &23.44
   \\ 
     Bilinear                   &31 &8 &14.74M     &-      &29.94    &27.66    &26.10 &27.05 &24.06
    \\
     Bicubic                     &126 &20 &51.61M    &-      &31.03    &28.24    &26.68 &27.76 &24.59
    \\
     Lanczos2                      &494 &22 &110.59M   &-      &31.08    &28.26    &26.69 &27.77 &24.60
    \\
     Lanczos3                     &914 &30  &165.89M   &-      &31.48    &28.50    &26.90 &28.06 &24.81
    \\
    LeRF-L                   &80   &9 &26.15M     &0.72MB      &31.74 &28.39 &27.08 &28.24 &25.09 \\
    LeRF-G                   &110  &14 &57.94M     &1.67MB      &\underline{32.25}    &\underline{29.09}    &\underline{27.59}  &\underline{28.84}   &\underline{25.51}\\
    LeRF-Net                   &1,490 &15  &60.71G     &4.21MB      &\textbf{33.51}    &\textbf{29.97}    &\textbf{28.38} &\textbf{29.77} &\textbf{26.11}  \\ 
                                    \midrule
MetaSR \cite{DBLP:conf/cvpr/HuM0WT019}              &10,260  &590    &1.68T    &85.59MB      &35.00 &\underline{31.27} &29.25  &- &-  \\
LIIF \cite{DBLP:conf/cvpr/ChenL021}               &67,080   &677   &2.54T    &255.76MB      &34.99 &31.26 &29.27  &- &-  \\                          
SRWarp  \cite{DBLP:conf/cvpr/SonL21}              &-   &617   &1.01T    &212.12MB      & OOM & OOM & OOM &\underline{31.04} &26.75 \\
LTEW  \cite{DBLP:conf/eccv/LeeCJ22}               &-   &757   &1.71T    &196.13MB      &\underline{35.04}    &\textbf{31.32}    &\textbf{29.31} &\textbf{31.10} &\textbf{26.92}                          \\
LeRF-Net++                   &5,161 & 130 &0.94T     &65.99MB      &\textbf{35.13}    &31.20    &\underline{29.30} &30.72 &\underline{26.79}  \\ 
                                     \bottomrule
    \end{tabular}%
    \begin{tablenotes}
        \item The efficiency metrics are evaluated on producing a $1280 \times 720$ HD image through $\times 4$ upsampling. For efficiency-orientated methods (interpolation, LeRF-L, LeRF-G, and LeRF-Net), the CPU running time is evaluated on a OnePlus 7 Pro smartphone with a Qualcomm Snapdragon 855 CPU, while for performance-orientated methods (DNNs and LeRF-Net++), that is evaluated on a desktop computer with an Intel Xeon Gold 6278C 2.60GHz CPU. 
        \item GPU running time is evaluated on an NVIDIA RTX 3090 GPU. Due to a CPU compatiblity issue of the custom CUDA operator in SRWarp and LTEW, only GPU running times are reported.
        \item OOM denotes out-of-memory runtime error when inference SRWarp. cPSNR denotes PNSR values averaged across three color channels. The best and second-best results are \textbf{highlighted} and \underline{underlined}. 
    \end{tablenotes}
\end{threeparttable}

    }
    \label{tab:comp}
    \end{table*}

%% file: tables/lpips_table.tex
\renewcommand{\arraystretch}{1.2}

\begin{table*}[htbp]
  \centering
  \caption{Quantitative comparison in LPIPS for arbitrary-scale upsampling. }
    \begin{threeparttable}
\resizebox{\textwidth}{!}{%
        
    \begin{tabular}{lcccccccccccccccccccc}
    \toprule
    \multirow{2}{*}{Method} &\multicolumn{3}{c}{Set5} &\multicolumn{3}{c}{Set14} &\multicolumn{3}{c}{BSDS100} &\multicolumn{3}{c}{Urban100} &\multicolumn{3}{c}{Manga109} \\ \cmidrule(lr){2-4} \cmidrule(lr){5-7} \cmidrule(lr){8-10} \cmidrule(lr){11-13} \cmidrule(lr){14-16} 
                           &$\times2$   &$\times3$   &$\times4$   
                        &$\times2$   &$\times3$   &$\times4$   
                        &$\times2$   &$\times3$   &$\times4$   
                        &$\times2$   &$\times3$   &$\times4$   
                        &$\times2$   &$\times3$   &$\times4$    \\ \midrule
                                    Bilinear   &0.1674	&0.2421	&0.3572	
                                            &0.2444	&0.3433	&0.4645
                                            &0.3181	&0.4228	&0.5503	
                                            &0.2640	&0.3764	&0.4965	
                                            &0.1502	&0.2403	&0.3481	
                                    \\
                                    Bicubic    &0.1261	&0.2508	&0.3461	
                                            &0.1958	&0.3603	&0.4573	
                                            &0.2643	&0.4451	&0.5432	
                                            &0.2203	&0.3947	&0.4932	
                                            &0.1107	&0.2416	&0.3340	   
                                    \\ 
                                    Lanczos3  &0.1349	&0.2527	&0.3501	
                                            &0.2177	&0.3649	&0.4628	
                                            &0.2968	&0.4538	&0.5530	
                                            &0.2476	&0.3953	&0.4940	
                                            &0.1209	&0.2389	&0.3320	
                                     \\ 
                                     RAISR* \cite{DBLP:journals/tci/RomanoIM17}  &0.0784	&0.1669	&0.2421	
                                             &0.1418	&0.2784	&0.3588	
                                             &0.1984	&0.3651	&0.4397	
                                             &0.1567	&0.3062	&0.3915	
                                             &0.0582	&0.1591	&0.2461	
                                      \\ 
                            LeRF-L       &0.0914	&0.1578	&0.2612	   
                            &0.1475	&0.2557	&0.3628	 
                            &0.2183	&0.3378	&0.4472	 
                            &0.1783	&0.2973	&0.3867	
                            &0.0788	&0.1691	&0.2620 
                            \\ 
                            LeRF-G       &\underline{0.0504}	&\underline{0.1062}	&\underline{0.1830}	   
                            &\underline{0.0879}	&\underline{0.2022}	&\underline{0.3023}	 
                            &\underline{0.1472}	&\underline{0.2906}	&\underline{0.3995}	 
                            &\underline{0.1096}	&\underline{0.2377}	&\underline{0.3374}	
                            &\underline{0.0325}	&\underline{0.0992}	&\underline{0.1699} 
                            \\ 
                                LeRF-Net       &\textbf{0.0499}	&\textbf{0.0987}	&\textbf{0.1665}	   
                                       &\textbf{0.0870}	&\textbf{0.1872}	&\textbf{0.2821}	 
                                       &\textbf{0.1332}	&\textbf{0.2654}	&\textbf{0.3741}	 
                                       &\textbf{0.0752}	&\textbf{0.1722}	&\textbf{0.2576}	
                                       &\textbf{0.0242}	&\textbf{0.0700}	&\textbf{0.1260} 
                                \\ 
                                \midrule
                            RCAN* \cite{DBLP:conf/eccv/ZhangLLWZF18}  &\underline{0.0544}	&0.1385	&0.1956	
                            &\underline{0.0879}	&0.2270	&0.3049	
                            &\underline{0.1383}	&0.3115	&0.3968	
                            &\textbf{0.0532}	&0.1859	&0.2687	
                            &\textbf{0.0215}	&0.0833	&0.1407	
                     \\ 
                                MetaSR \cite{DBLP:conf/cvpr/HuM0WT019}  &0.0576	&0.1213	&\textbf{0.1706}	
                                        &0.0952	&0.2083	&0.2886	
                                        &0.1490	&0.2857	&0.3766	
                                        &0.0599	&0.1452	&0.2179	
                                        &0.0236	&0.0643	&0.1052	
                                 \\ 
                                 LIIF \cite{DBLP:conf/cvpr/ChenL021}   &0.0583	&0.1223	&\underline{0.1714}	
                                         &0.0953	&0.2090	&0.2918	
                                         &0.1557	&0.2907	&0.3793	
                                         &0.0622	&0.1470	&0.2216	
                                         &0.0246	&0.0656	&0.1083	
                                  \\  
                                  SRWarp  \cite{DBLP:conf/cvpr/SonL21}   &\textbf{0.0538}	&\underline{0.1201}	&0.1689	
                                          &\textbf{0.0871}	&\textbf{0.1979}	&\textbf{0.2760}	
                                          &\textbf{0.1365}	&\textbf{0.2689}	&\textbf{0.3546}	
                                          &0.0543	&\textbf{0.1366}	&\textbf{0.2052}	
                                          &0.0218	&\textbf{0.0611}	&\textbf{0.1018}	
                                   \\ 
                                   LTEW  \cite{DBLP:conf/eccv/LeeCJ22}   &0.0568	&0.1230	&0.1725	
                                           &0.0909	&0.2025	&\underline{0.2824}	
                                           &0.1443	&0.2756	&0.3687	
                                           &0.0574	&\underline{0.1392}	&\underline{0.2089}	
                                           &0.0232	&0.0628	&0.1046	
                                    \\ 
                                    LeRF-Net++  &\underline{0.0544}	&\textbf{0.1198}	&0.1715	
                                            &\underline{0.0879}	&\underline{0.2021}	&0.2830	
                                            &\underline{0.1383}	&0.2789	&0.3711	
                                            &\textbf{0.0532}	&\underline{0.1392}	&0.2180	
                                            &\textbf{0.0215}	&\underline{0.0625}	&\underline{0.1036}	
                                    \\ \bottomrule
    \end{tabular}%
    }

    \begin{tablenotes}
    \item * denotes that we combine fixed-scale super-resolution methods with Bicubic interpolation to achieve arbitrary-scale upsampling. 
    \item Lower is better. The best and second-best results are \textbf{highlighted} and \underline{underlined}. 
\end{tablenotes}

\end{threeparttable}
    \label{tab:lpips}
    \end{table*}

%% file: figures/fig_continuous.tex
\begin{figure*}[t]
    \centering
    \includegraphics[width=\textwidth]{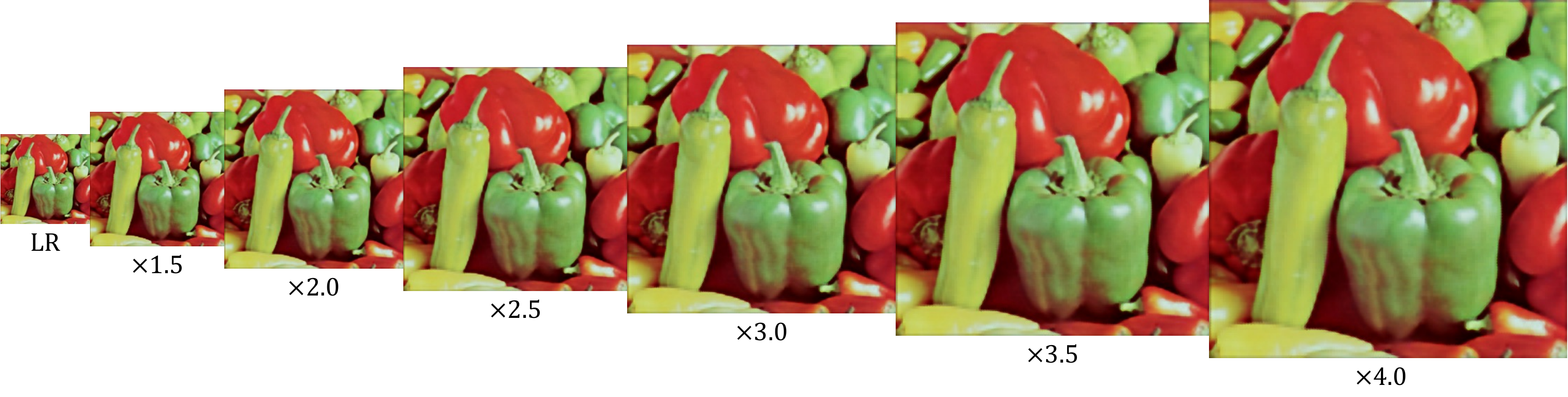}
    \caption{Continuous upsampling results of LeRF. More continuous resampling results are available on our project page at $\text{https://lerf.pages.dev}$.} 
    \label{fig:continuous}
\end{figure*}

%% file: figures/visual_rule.tex

\renewcommand{\vsp}{-6mm}

%
\begin{figure}[t]
	\scriptsize
%
%

\renewcommand{\h}{0.09}
\renewcommand{\w}{0.09}
\renewcommand{\hrs}{0.25}

\hspace{-10mm}

\renewcommand{\name}{visual/rule-based/X2.00_2.00/Set14-comic/Y100_X100/comic-}
\setlength{\g}{-4mm}
\begin{tabular}{cc}
	\hspace{-4mm}
	\begin{adjustbox}{valign=t}
		\begin{tabular}{c}
			\includegraphics[height=\h \textwidth, width=\h \textwidth]{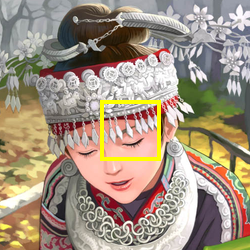}
			\\ HR
		\end{tabular}

\vspace{\vsp}	\end{adjustbox}
	\hspace{-4mm}
	\begin{adjustbox}{valign=t}
		\begin{tabular}{ccccccc}
			\includegraphics[height=\h \textwidth, width=\w \textwidth]{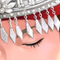}~\hspace{\g} &
			\includegraphics[height=\h \textwidth, width=\w \textwidth]{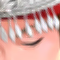}~\hspace{\g} & 
			\includegraphics[height=\h \textwidth, width=\w \textwidth]{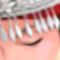}~\hspace{\g} & 
			\includegraphics[height=\h \textwidth, width=\w \textwidth]{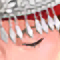}
			\\
			GT \hspace{\g} &
			NEDI \hspace{\g} &
			SKR \hspace{\g} &
			LeRF-G
		\end{tabular}

\vspace{\vsp}	\end{adjustbox}
\end{tabular}

\renewcommand{\name}{visual/rule-based/X2.00_2.00/Set14-ppt3/Y40_X230/ppt3-}
\setlength{\g}{-4mm}
\begin{tabular}{cc}
	\hspace{-4mm}
	\begin{adjustbox}{valign=t}
		\begin{tabular}{c}
			\includegraphics[height=\h \textwidth, width=\h \textwidth]{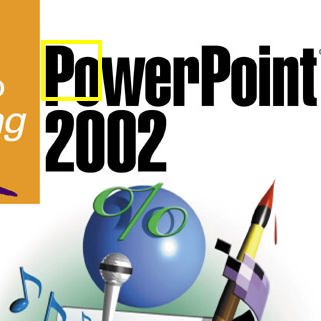}
			\\ HR
		\end{tabular}

\vspace{\vsp}	\end{adjustbox}
	\hspace{-4mm}
	\begin{adjustbox}{valign=t}
		\begin{tabular}{ccccccc}
			\includegraphics[height=\h \textwidth, width=\w \textwidth]{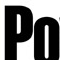}~\hspace{\g} &
			\includegraphics[height=\h \textwidth, width=\w \textwidth]{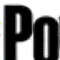}~\hspace{\g} & 
			\includegraphics[height=\h \textwidth, width=\w \textwidth]{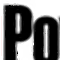}~\hspace{\g} & 
			\includegraphics[height=\h \textwidth, width=\w \textwidth]{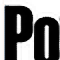}
			\\
			GT \hspace{\g} &
			NEDI \hspace{\g} &
			SKR \hspace{\g} &
			LeRF-G
		\end{tabular}

\vspace{\vsp}	\end{adjustbox}
\end{tabular}

\renewcommand{\name}{visual/rule-based/X2.00_2.00/Manga109-KyokugenCyclone/Y560_X410/KyokugenCyclone-}
\setlength{\g}{-4mm}
\begin{tabular}{cc}
	\hspace{-4mm}
	\begin{adjustbox}{valign=t}
		\begin{tabular}{c}
			\includegraphics[height=\h \textwidth, width=\h \textwidth]{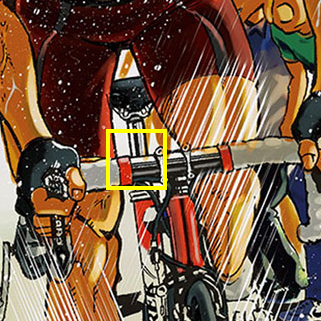}
			\\ HR
		\end{tabular}

\vspace{\vsp}	\end{adjustbox}
	\hspace{-4mm}
	\begin{adjustbox}{valign=t}
		\begin{tabular}{ccccccc}
			\includegraphics[height=\h \textwidth, width=\w \textwidth]{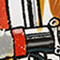}~\hspace{\g} &
			\includegraphics[height=\h \textwidth, width=\w \textwidth]{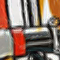}~\hspace{\g} & 
			\includegraphics[height=\h \textwidth, width=\w \textwidth]{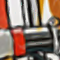}~\hspace{\g} & 
			\includegraphics[height=\h \textwidth, width=\w \textwidth]{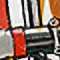}
			\\
			GT \hspace{\g} &
			NEDI \hspace{\g} &
			SKR \hspace{\g} &
			LeRF-G
		\end{tabular}

\vspace{\vsp}	\end{adjustbox}
\end{tabular}

\caption{Qualitative comparison with rule-based adaptive resampling methods for $\times2$ upsampling. We also include Bicubic as a companion. From top to bottom, the example images are: \textit{comic} (Set14), \textit{ppt3} (Set14), and \textit{KyokugenCyclone} (Manga109). Best view on screen and in color.}
\label{fig:visual_rule}
\end{figure}

%% file: figures/visual_warp.tex
\begin{figure*}[t]
    \centering
    \includegraphics[width=\textwidth]{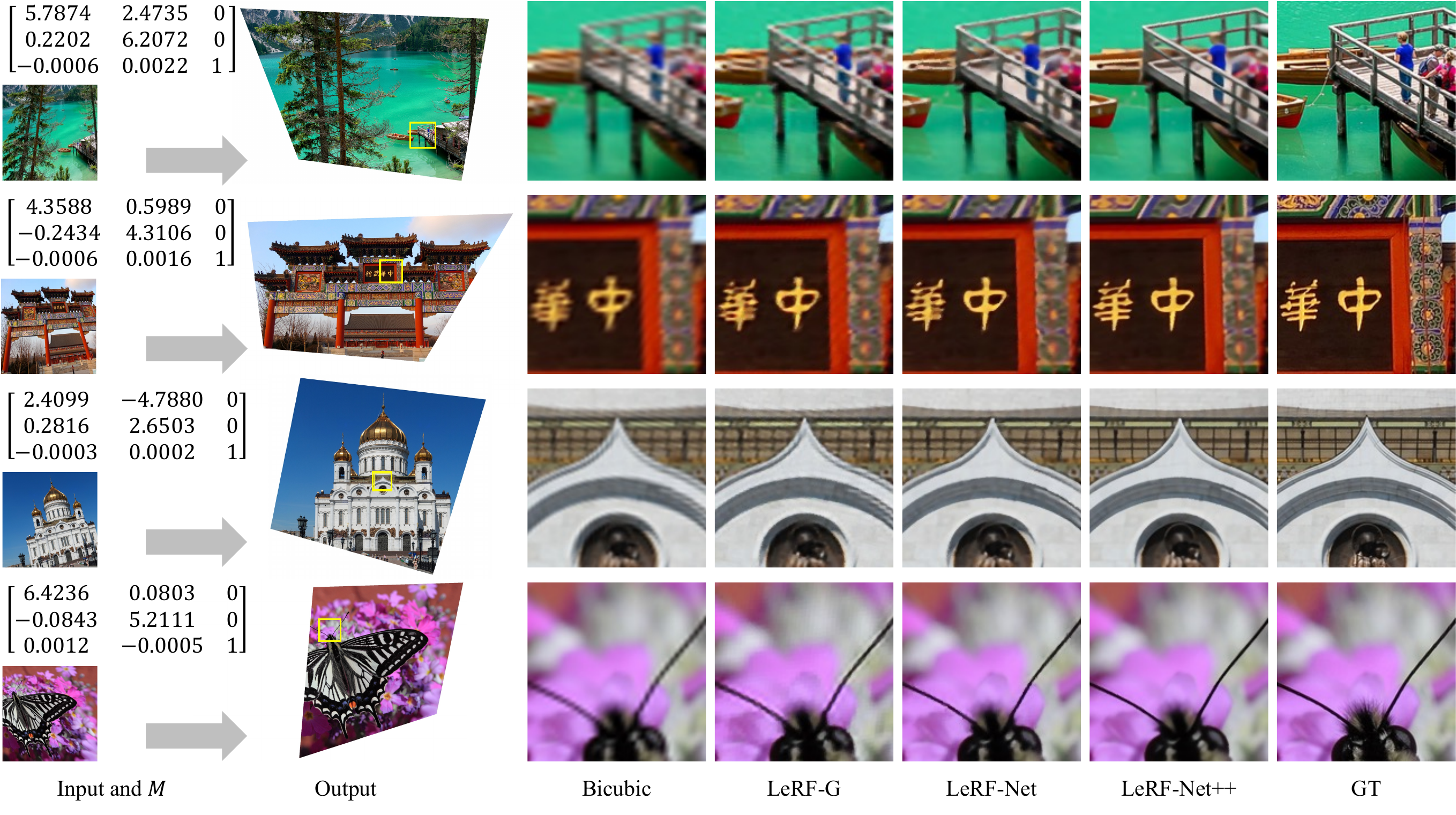}
    \caption{Qualitative comparison for homographic warping. The homographic transformation matrix $M$ is shown along the input image. From top to bottom, the example images are: \emph{0807} (out-scale), \emph{0826} (out-scale), and \emph{0882} (out-scale). Best view in color and on screen.} 
    \label{fig:visual_warp}
\end{figure*}

%% file: figures/fig_tradeoff.tex
\begin{figure}[t]
    \centering
    \includegraphics[width=\columnwidth]{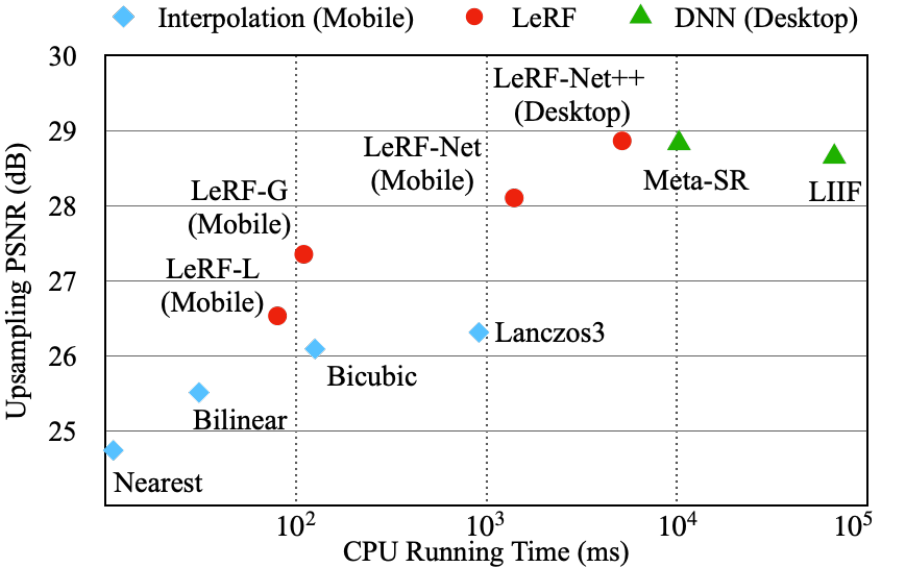}
    \caption{Performance-efficiency trade-off of arbitrary-scale upsampling methods. PSNR values are obtained on Set14 for $\times4$ upsampling. The running time is evaluated on mobile and desktop CPUs for producing $1280 \times 720$ images through $\times 4$ upsampling.} 
    \label{fig:tradeoff_cpu}
\end{figure}

%% file: figures/visual_affine.tex
\begin{figure}[t]
    \centering
    \includegraphics[width=\columnwidth]{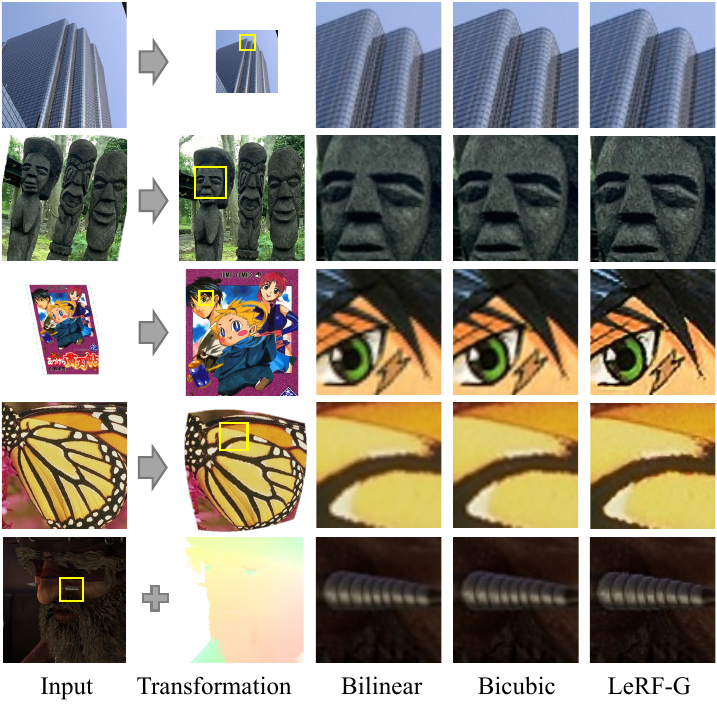}
    \caption{Visual comparison of LeRF with interpolation methods under homographic transformations (downsampling, rotation, and sheering) and arbitrary warping (according to a barrel-shaped distortion and optical flow). } 
    \label{fig:visual_affine}
\end{figure}

%% file: figures/visual_down.tex
\begin{figure}[t]
    \centering
    \includegraphics[width=\columnwidth]{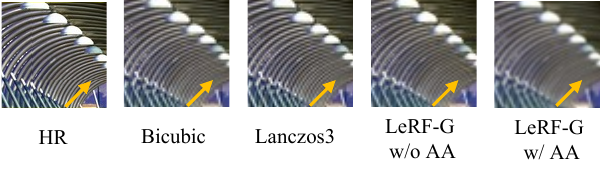}
    \caption{LeRF-G with AA filter reduces aliasing artifacts for image downsampling in regions with highly dense textures.} 
    \label{fig:visual_down}
\end{figure}

%% file: figures/fig_hyper_vis.tex
\begin{figure}[t]
    \centering
    \includegraphics[width=\columnwidth]{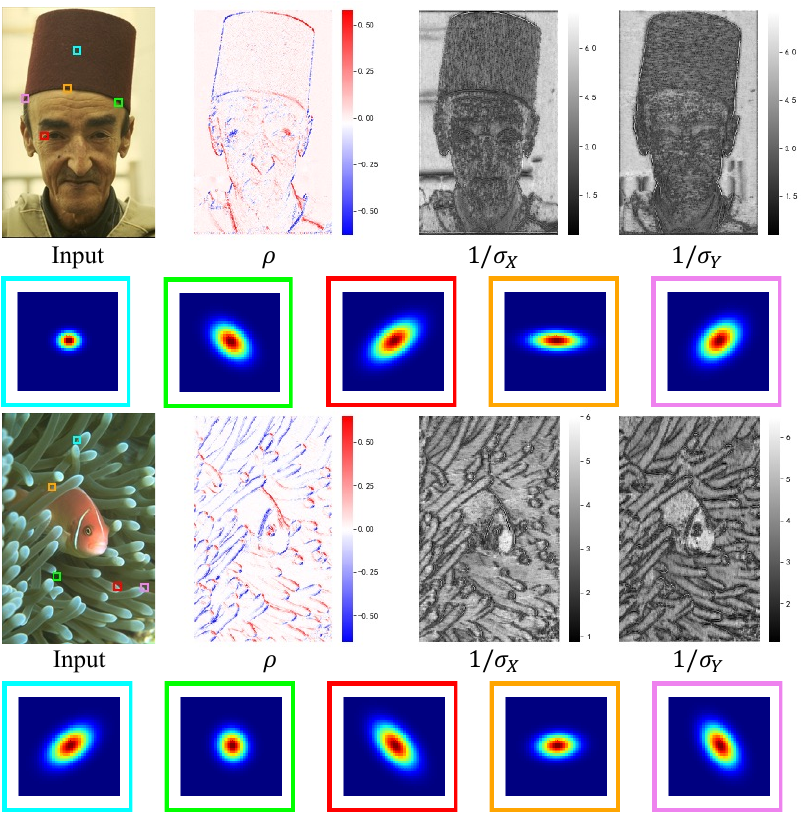}
    \caption{Visualization of the pixel-wise hyper-parameters and the corresponding resampling functions. The shapes of the predicted resampling functions are well adapted to local structures.} 
    \label{fig:hyper_vis}
\end{figure}

%% file: figures/fig_optim_kernel.tex
\begin{figure}[t]
    \centering
    \includegraphics[width=\columnwidth]{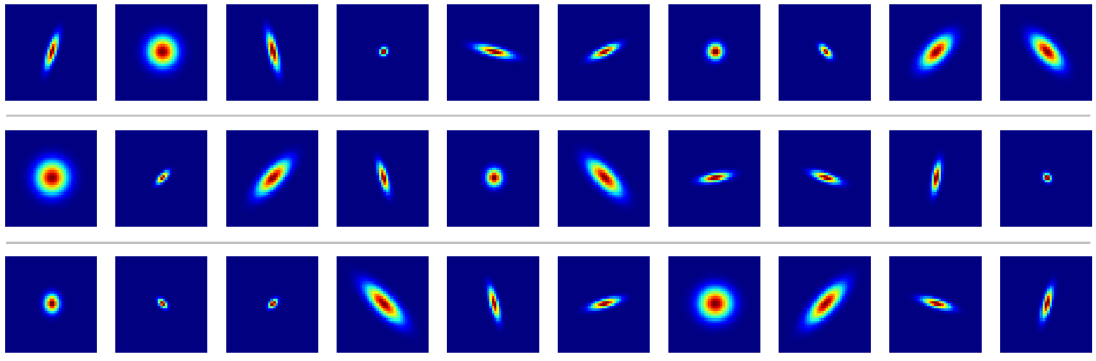}
    \caption{Visualization of GT-optimized resampling function clustering centroids on Set14 for $\times2$, $\times3$, and $\times4$ upsampling, respectively. They show consistency across different upsampling scales.} 
    \label{fig:optim_kernel}
\end{figure}

%% file: tables/abl_table1.tex

    \renewcommand{\arraystretch}{1}
    \begin{table}[t]
      \caption{Ablation experiments on Set5 on the the design of the adaptive resampling function.}
        \resizebox{\columnwidth}{!}{%
        \begin{tabular}{lrrHH|ccc} \toprule
          & MACs & Storage & Hyper RF & Supp. Size & $\times2$ & $\times3$ & $\times4$ \\ \midrule
          fixed Gaussian $\Phi^G_{(0, 1, 1)}$ & 45.04M  &244.69KB &$3 \times 3$ &$4 \times 4$ & 30.75 & 27.70 & 26.31 \\  
          isotropic Keys $\mathtt{cubic}$ & 126.94M  &734.07KB &$3 \times 3$ &$4 \times 4$ & 34.68 & 31.17 & 29.08 \\  \midrule
          LeRF-L & 26.15M  &734.07KB &$3 \times 3$ &$4 \times 4$ & 34.84 & 30.72 & 29.13 \\ 
          LeRF-G & 57.94M  &1.67MB &$3 \times 3$ &$4 \times 4$ & 35.71 & 32.02 & 30.15 \\
          \bottomrule     
        \end{tabular}%
        }
        \label{tab:abl1}
        \end{table}

%% file: figures/visual_abl1.tex
\begin{figure}[t]
    \centering
    \includegraphics[width=\columnwidth]{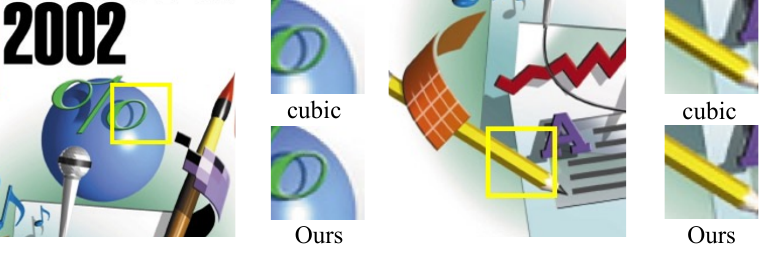}
    \caption{The non-steerable Keys $\mathtt{cubic}$ function produces artifacts along diagonal edges.} 
    \label{fig:visual_abl1}
\end{figure}

%% file: tables/abl_table3.tex

\renewcommand{\arraystretch}{1}
\begin{table}[t]
  \caption{Ablation experiments on Set5 on the LUT acceleration and our adaptations.}
    \resizebox{\columnwidth}{!}{%
    \begin{tabular}{lrrHH|ccc} \toprule
      & MACs & Storage & Hyper RF & Supp. Size & $\times2$ & $\times3$ & $\times4$ \\ \midrule
      LeRF-G & 57.94M  &1.67MB &$3 \times 3$ &$4 \times 4$ & 35.71 & 32.02 & 30.15 \\ \midrule
      w/o LUT acceleration & 21.75G  &1.59MB &$3 \times 3$ &$5 \times 5$ & 36.11 & 32.23 & 30.28 \\ 
      w/o DE & 57.94M  &978.76KB &$3 \times 3$ &$4 \times 4$ & 34.44 & 31.50 & 29.47 \\ 
      w/o ``CX'' pattern & 49.65M  &326.25KB &$3 \times 3$ &$4 \times 4$ & 34.90 & 31.79 & 29.70 \\
      w/o pre-processing & 53.45M  &1.43MB &$3 \times 3$ &$4 \times 4$ & 34.43 & 31.41 & 29.33 \\
      \bottomrule     
    \end{tabular}%
    }
    \label{tab:abl3}
    \end{table}

%% file: figures/visual_abl2.tex
\begin{figure}[t]
    \centering
    \includegraphics[width=\columnwidth]{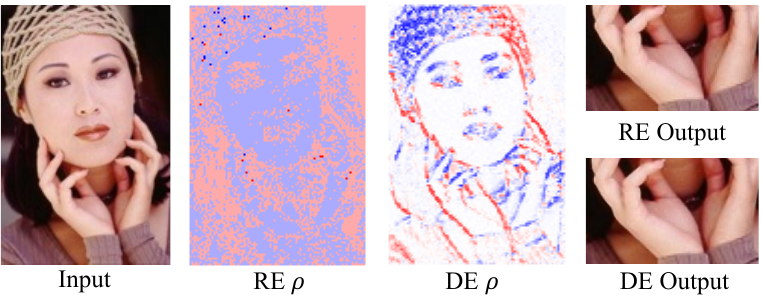}
    \caption{Replacing RE with DE enables the learning of edge orientations.} 
    \label{fig:visual_abl2}
\end{figure}

%% file: tables/abl_net_table.tex

\renewcommand{\arraystretch}{1}
\begin{table}[t]
  \caption{Ablation analysis on the network architeture on the Set5 dataset.}
  \resizebox{\columnwidth}{!}{%
  \begin{threeparttable}
    
    \begin{tabular}{lHrr|ccc} \toprule
      & \# of Params.  & MACs & RF & $\times2$ & $\times3$ & $\times4$ \\ \midrule
      LeRF-G & - & 57.69M & $7\times7$ & 35.71 & 32.02 & 30.15 \\
      LeRF-G (w/o LUT acc.) & 84.36K & 21.75G & $7\times7$ & 36.11 & 32.23 & 30.28 \\
      LeRF-Net (conv. only) & 79.24K & 4.56G & $13\times13$ & 36.23 & 32.58 & 30.71 \\
      LeRF-Net (nf.=16) & 276.48K & 3.97G & $43\times43$ &36.44 &32.77 & 30.94 \\ 
      LeRF-Net (nf.=32) & 1.38K & 15.40G & $43\times43$ &36.46 &32.87 & 31.11 \\ 
      LeRF-Net (nf.=64, Ours) & 4.21MB & 60.71G & $43\times43$ &36.57 &33.04 & 31.26 \\ 
      LeRF-Net (nf.=80) & 6.56M & 94.59G & $43\times43$ &36.60 &33.07 & 31.28 \\ 
      LeRF-Net (nf.=96) & 7.60M & 109.38G & $43\times43$ &36.52 &33.06 & 31.30 \\ 
      \bottomrule     
    \end{tabular}%
    \begin{tablenotes}
      \item LUT acc. denotes the LUT acceleration of the multi-brance DNN in LeRF-G, conv. the convolutional layer, and nf. the channel number of convolutional layers in LeRF-Net.
  \end{tablenotes}    
  \end{threeparttable}
    }
    \label{tab:abl_net}
    \end{table}

%% file: figures/visual_abl_pre.tex
\begin{figure}[t]
    \centering
    \includegraphics[width=\columnwidth]{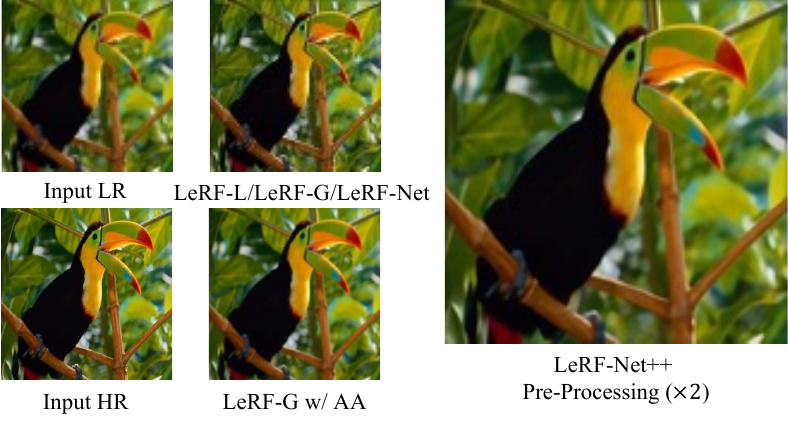}
    \caption{The pre-processing stage plays the role of enhancing edges (LeRF-L, LeRF-G, and LeRF-Net), pre-upsampling (LeRF-Net++), and reducing high-frequency details for downsampling (LeRF-G w/ AA).} 
    \label{fig:visual_abl_pre}
\end{figure}

%% file: tables/abl_table2.tex

\renewcommand{\arraystretch}{1}
\begin{table}[t]
  \centering
  \caption{Ablation experiments on Set5 on data pairs for training.}
    \resizebox{0.95\columnwidth}{!}{%
    \begin{tabular}{lrrHH|ccc} \toprule
      & MACs & Storage & Hyper RF & Supp. Size & $\times2$ & $\times3$ & $\times4$ \\ \midrule
      LeRF-G ($\times 4$ data) & 57.94M  &1.67MB &$3 \times 3$ &$4 \times 4$ & 35.71 & 32.02 & 30.15 \\ \midrule
      $\times 2$ data training & 57.94M  &1.67MB &$3 \times 3$ &$4 \times 4$ & 35.67 & 32.11 & 29.91 \\  
      mixed-scale training & 57.94M  &1.67MB &$3 \times 3$ &$4 \times 4$ & 35.67 & 32.26 & 30.06 \\
      \bottomrule     
    \end{tabular}%
    }
    \label{tab:abl2}
    \end{table}

%% file: tables/comparison_lut.tex
\renewcommand{\arraystretch}{1}
\begin{table}[t]
  \caption{Additional comparison with LUT-based super-resolution methods on the Set5 dataset. }
  \centering
  \resizebox{\columnwidth}{!}{%
  \begin{threeparttable}
    \begin{tabular}{lHrr|ccc} \toprule
      & \# of Params.  & MACs & RunTime & $\times2$ & $\times3$ & $\times4$ \\ \midrule
      Bilinear & - & 14.74M & 31 & 32.23 & 29.53 & 27.55 \\
      Bicubic & - & 51.61M & 126 & 33.64 & 30.39 & 28.42 \\
      $\text{SR-LUT}^{\sharp}$ & - & 53.33M & 137 & 34.47 & 31.00 & 29.00 \\
      LeRF-G  & - & 57.94M & 110 & 35.71 & 32.02 & 30.15 \\ 
      \midrule
      SR-LUT* & - & 57.66M & 149 & 35.53 & 31.91 & 29.75 \\ 
      LeRF-G w/ SR-LUT & - & 59.45M & 101 & 35.53 & 31.92 & 29.78  \\
      MuLUT* & - & 73.26M & 174 & 36.65 & 32.47 & 30.25 \\ 
      LeRF-G w/ MuLUT & - & 75.05M & 140 & 36.65 & 32.43 & 30.28 \\ 
      \bottomrule     
    \end{tabular}%
    \begin{tablenotes}
      \item The efficiency metrics are evaluated on mobile CPU for producing a $1280 \times 720$ HD image through $\times 4$ upsampling. $\text{SR-LUT}^{\sharp}$ denotes SR-LUT is retrained with Bicubic on mixed-scale LR-HR data pairs. * denotes that we combine fixed-scale LUT-based super-resolution methods ($\times 2$ upsampling) with Bicubic to achieve arbitrary-scale upsampling. 
  \end{tablenotes}    
  \end{threeparttable}

}
\label{tab:comparison_lut}
    \end{table}

%% file: parts/conclusion.tex
\section{Discussion and Conclusion}

\noindent\textbf{Open questions.}
1) From a higher point of view, LeRF represents an opposite direction in modeling the resampling process, compared to existing DNN-based methods that utilize a kernel prediction network for predicting resampling weights. This line of work assumes resampling weights is only dependent on coordinate offsets that is ultimately determined by the geometric transformation for resampling. On the other hand, LeRF takes the other side by learning resampling functions only based on the local image content, thus our learning process is agnostic to the geometric transformation (see TABLE~\ref{tab:abl2}). Interestingly, as shown in Fig.~\ref{fig:visual_affine} and Fig.~\ref{fig:optim_kernel}, LeRF shows great generalization capacity even though ALL transformations but upsampling are unseen during the construction process of LeRF models. Thus, we argue that there is a chance to incorporate these two factors, \emph{i.e.}, geometric transformation and image content, into the resampling process like bilateral filtering \cite{DBLP:conf/dagm/AurichW95,DBLP:conf/iccv/TomasiM98,DBLP:journals/ftcgv/KornprobstTD09} in the field of image filtering, which we leave as our future work.
2) Another open question is the usage of LeRF in non-grid data like point clouds. Recently, 3D Gaussians emerge as an efficient solution for modeling 3D content \cite{DBLP:journals/tog/KerblKLD23}. LeRF shares a similar idea of representing a local intensity surface with parameterized explicit functions. In this sense, LeRF has the potential to be integrated as a tool for enhancing the 3D representations at high efficiency.

\noindent\textbf{Conclusion remarks.}
To summarize, in this work, we propose LeRF, a novel method for image resampling by integrating learned structural priors into the adaptive resampling function. Furthermore, with LUT acceleration, our efficiency-orientated models play as a superior competitor to widely used interpolation methods. We show its superior performance, high efficiency, and versatility for arbitrary transformations. We also propose a performance-orientated LeRF model for empowering existing fixed-scale upsampling methods to achieve image resampling with similar performance as tailored solutions. Our method reveals the power of a novel combination of classic image processing techniques and the recent learning-from-data paradigm. We hope this work could increase the attention of the research community to this promising direction.

%% file: lerf.bbl
\begin{thebibliography}{10}
\providecommand{\url}[1]{#1}
\csname url@samestyle\endcsname
\providecommand{\newblock}{\relax}
\providecommand{\bibinfo}[2]{#2}
\providecommand{\BIBentrySTDinterwordspacing}{\spaceskip=0pt\relax}
\providecommand{\BIBentryALTinterwordstretchfactor}{4}
\providecommand{\BIBentryALTinterwordspacing}{\spaceskip=\fontdimen2\font plus
\BIBentryALTinterwordstretchfactor\fontdimen3\font minus \fontdimen4\font\relax}
\providecommand{\BIBforeignlanguage}[2]{{%
\expandafter\ifx\csname l@#1\endcsname\relax
\typeout{** WARNING: IEEEtran.bst: No hyphenation pattern has been}%
\typeout{** loaded for the language `#1'. Using the pattern for}%
\typeout{** the default language instead.}%
\else
\language=\csname l@#1\endcsname
\fi
#2}}
\providecommand{\BIBdecl}{\relax}
\BIBdecl

\bibitem{dodgson1992image}
N.~A. Dodgson, ``Image resampling,'' University of Cambridge, Computer Laboratory, Tech. Rep., 1992.

\bibitem{Gardner:48}
I.~C. Gardner and F.~E. Washer, ``Lenses of extremely wide angle for airplane mapping,'' \emph{J. Opt. Soc. Am.}, vol.~38, no.~5, pp. 421--431, 1948.

\bibitem{mrak2016high}
M.~Mrak, M.~Grgic, and M.~Kunt, \emph{High-Quality Visual Experience: Creation, Processing and Interactivity of High-Resolution and High-Dimensional Video Signals}, ser. Signals and Communication Technology.\hskip 1em plus 0.5em minus 0.4em\relax Springer Berlin Heidelberg, 2016.

\bibitem{smith1986industrial}
T.~G. Smith and G.~Lucas, \emph{Industrial Light \& Magic: The Art of Special Effects}.\hskip 1em plus 0.5em minus 0.4em\relax Virgin, 1986.

\bibitem{DBLP:conf/eccv/DongLHT14}
C.~Dong, C.~C. Loy, K.~He, and X.~Tang, ``Learning a deep convolutional network for image super-resolution,'' in \emph{ECCV}, 2014.

\bibitem{DBLP:conf/cvpr/KimLL16a}
J.~Kim, J.~K. Lee, and K.~M. Lee, ``Accurate image super-resolution using very deep convolutional networks,'' in \emph{CVPR}, 2016.

\bibitem{DBLP:conf/cvpr/LimSKNL17}
B.~Lim, S.~Son, H.~Kim, S.~Nah, and K.~M. Lee, ``Enhanced deep residual networks for single image super-resolution,'' in \emph{CVPR Workshops}, 2017.

\bibitem{DBLP:conf/cvpr/HuM0WT019}
X.~Hu, H.~Mu, X.~Zhang, Z.~Wang, T.~Tan, and J.~Sun, ``{Meta-SR}: {A} magnification-arbitrary network for super-resolution,'' in \emph{CVPR}, 2019.

\bibitem{DBLP:journals/tip/SunC20}
W.~Sun and Z.~Chen, ``Learned image downscaling for upscaling using content adaptive resampler,'' \emph{{IEEE} Trans. Image Process.}, vol.~29, pp. 4027--4040, 2020.

\bibitem{DBLP:conf/cvpr/SonL21}
S.~Son and K.~M. Lee, ``{SRWarp}: Generalized image super-resolution under arbitrary transformation,'' in \emph{CVPR}, 2021.

\bibitem{Keys1981CubicCI}
R.~G. Keys, ``Cubic convolution interpolation for digital image processing,'' \emph{IEEE Trans. Acoust.}, vol.~29, pp. 1153--1160, 1981.

\bibitem{DBLP:conf/cvpr/ChenL021}
Y.~Chen, S.~Liu, and X.~Wang, ``Learning continuous image representation with local implicit image function,'' in \emph{CVPR}, 2021.

\bibitem{DBLP:conf/iccv/Wang0L0AG21}
L.~Wang, Y.~Wang, Z.~Lin, J.~Yang, W.~An, and Y.~Guo, ``Learning {A} single network for scale-arbitrary super-resolution,'' in \emph{ICCV}, 2021.

\bibitem{DBLP:conf/nips/YangSYL21}
J.~Yang, S.~Shen, H.~Yue, and K.~Li, ``Implicit transformer network for screen content image continuous super-resolution,'' in \emph{NeurIPS}, 2021.

\bibitem{DBLP:conf/cvpr/JoK21}
Y.~Jo and S.~J. Kim, ``Practical single-image super-resolution using look-up table,'' in \emph{CVPR}, 2021.

\bibitem{splut}
C.~Ma, J.~Zhang, J.~Zhou, and J.~Lu, ``Learning series-parallel lookup tables for efficient image super-resolution,'' in \emph{ECCV}, 2022.

\bibitem{mulut}
J.~Li, C.~Chen, Z.~Cheng, and Z.~Xiong, ``{MuLUT}: Cooperating multiple look-up tables for efficient image super-resolution,'' in \emph{ECCV}, 2022.

\bibitem{mulut23}
------, ``Toward {DNN} of {LUTs}: Learning efficient image restoration with multiple look-up tables,'' \emph{arxiv}, vol. 2303.14506, 2023.

\bibitem{lerf_cvpr}
J.~Li, C.~Chen, W.~Huang, Z.~Lang, F.~Song, Y.~Yan, and Z.~Xiong, ``Learning steerable function for efficient image resampling,'' in \emph{CVPR}, 2023.

\bibitem{DBLP:conf/siggraph/MitchellN88}
D.~P. Mitchell and A.~N. Netravali, ``Reconstruction filters in computer-graphics,'' in \emph{SIGGRAPH}, 1988.

\bibitem{DBLP:conf/icip/AllebachW96}
J.~P. Allebach and P.~W. Wong, ``Edge-directed interpolation,'' in \emph{ICIP}, 1996.

\bibitem{DBLP:journals/tip/LiO01a}
X.~Li and M.~T. Orchard, ``New edge-directed interpolation,'' \emph{{IEEE} Trans. Image Process.}, vol.~10, no.~10, pp. 1521--1527, 2001.

\bibitem{DBLP:journals/tip/ZhangW06a}
L.~Zhang and X.~Wu, ``An edge-guided image interpolation algorithm via directional filtering and data fusion,'' \emph{{IEEE} Trans. Image Process.}, vol.~15, no.~8, pp. 2226--2238, 2006.

\bibitem{DBLP:journals/tip/WangW07}
Q.~Wang and R.~K. Ward, ``A new orientation-adaptive interpolation method,'' \emph{{IEEE} Trans. Image Process.}, vol.~16, no.~4, pp. 889--900, 2007.

\bibitem{DBLP:journals/tip/TakedaFM07}
H.~Takeda, S.~Farsiu, and P.~Milanfar, ``Kernel regression for image processing and reconstruction,'' \emph{{IEEE} Trans. Image Process.}, vol.~16, no.~2, pp. 349--366, 2007.

\bibitem{DBLP:journals/tip/LeeY10}
Y.~J. Lee and J.~Yoon, ``Nonlinear image upsampling method based on radial basis function interpolation,'' \emph{{IEEE} Trans. Image Process.}, vol.~19, no.~10, pp. 2682--2692, 2010.

\bibitem{DBLP:journals/tip/ZhangGTL12}
K.~Zhang, X.~Gao, D.~Tao, and X.~Li, ``Single image super-resolution with non-local means and steering kernel regression,'' \emph{{IEEE} Trans. Image Process.}, vol.~21, no.~11, pp. 4544--4556, 2012.

\bibitem{DBLP:journals/tci/RomanoIM17}
Y.~Romano, J.~Isidoro, and P.~Milanfar, ``{RAISR:} rapid and accurate image super resolution,'' \emph{{IEEE} Trans. Computational Imaging}, vol.~3, no.~1, pp. 110--125, 2017.

\bibitem{DBLP:conf/iccp/GetreuerGICOM18}
P.~Getreuer, I.~Garcia{-}Dorado, J.~Isidoro, S.~Choi, F.~Ong, and P.~Milanfar, ``{BLADE:} filter learning for general purpose computational photography,'' in \emph{ICCP}, 2018.

\bibitem{DBLP:journals/corr/abs-1712-06463}
X.~Jia, H.~Chang, and T.~Tuytelaars, ``Super-resolution with deep adaptive image resampling,'' \emph{arxiv}, vol. 1712.06463, 2017.

\bibitem{DBLP:conf/iccvw/ChenTWX17}
C.~Chen, X.~Tian, F.~Wu, and Z.~Xiong, ``{UDNet}: Up-down network for compact and efficient feature representation in image super-resolution,'' in \emph{ICCV Workshops}, 2017.

\bibitem{DBLP:conf/eccv/ZhangLLWZF18}
Y.~Zhang, K.~Li, K.~Li, L.~Wang, B.~Zhong, and Y.~Fu, ``Image super-resolution using very deep residual channel attention networks,'' in \emph{ECCV}, 2018.

\bibitem{DBLP:conf/cvpr/ChenXTZW19}
C.~Chen, Z.~Xiong, X.~Tian, Z.~Zha, and F.~Wu, ``Camera lens super-resolution,'' in \emph{CVPR}, 2019.

\bibitem{DBLP:journals/tog/WronskiGEKKLLM19}
B.~Wronski, I.~Garcia{-}Dorado, M.~Ernst, D.~Kelly, M.~Krainin, C.~Liang, M.~Levoy, and P.~Milanfar, ``Handheld multi-frame super-resolution,'' \emph{{ACM} Trans. Graph.}, vol.~38, no.~4, pp. 28:1--28:18, 2019.

\bibitem{DBLP:conf/iccvw/LiangCSZGT21}
J.~Liang, J.~Cao, G.~Sun, K.~Zhang, L.~V. Gool, and R.~Timofte, ``{SwinIR}: Image restoration using swin transformer,'' in \emph{ICCV Workshops}, 2021.

\bibitem{DBLP:conf/cvpr/XiaoFHCX21}
Z.~Xiao, X.~Fu, J.~Huang, Z.~Cheng, and Z.~Xiong, ``Space-time distillation for video super-resolution,'' in \emph{CVPR}, 2021.

\bibitem{pan2022towards}
Z.~Pan, B.~Li, D.~He, M.~Yao, W.~Wu, T.~Lin, X.~Li, and E.~Ding, ``Towards bidirectional arbitrary image rescaling: Joint optimization and cycle idempotence,'' in \emph{CVPR}, 2022.

\bibitem{yao2023bidirectional}
M.~Yao, D.~He, X.~Li, Z.~Pan, and Z.~Xiong, ``Bidirectional translation between uhd-hdr and hd-sdr videos,'' \emph{IEEE Trans. Multimedia}, 2023.

\bibitem{DBLP:conf/iccv/TalebiM21}
H.~Talebi and P.~Milanfar, ``Learning to resize images for computer vision tasks,'' in \emph{ICCV}, 2021.

\bibitem{DBLP:conf/cvpr/MildenhallBCSNC18}
B.~Mildenhall, J.~T. Barron, J.~Chen, D.~Sharlet, R.~Ng, and R.~Carroll, ``Burst denoising with kernel prediction networks,'' in \emph{CVPR}, 2018.

\bibitem{DBLP:conf/cvpr/Liang0GGT21}
J.~Liang, K.~Zhang, S.~Gu, L.~V. Gool, and R.~Timofte, ``Flow-based kernel prior with application to blind super-resolution,'' in \emph{CVPR}, 2021.

\bibitem{DBLP:conf/cvpr/0004CNWTL23}
X.~Wang, X.~Chen, B.~Ni, H.~Wang, Z.~Tong, and Y.~Liu, ``Deep arbitrary-scale image super-resolution via scale-equivariance pursuit,'' in \emph{CVPR}, 2023.

\bibitem{DBLP:conf/cvpr/BernasconiDSGS23}
M.~Bernasconi, A.~Djelouah, F.~Salehi, M.~H. Gross, and C.~Schroers, ``Kernel aware resampler,'' in \emph{CVPR}, 2023.

\bibitem{DBLP:conf/cvpr/WeiZ23}
M.~Wei and X.~Zhang, ``Super-resolution neural operator,'' in \emph{{IEEE/CVF} Conference on Computer Vision and Pattern Recognition, {CVPR} 2023, Vancouver, BC, Canada, June 17-24, 2023}, 2023, p. CVPR.

\bibitem{DBLP:conf/cvpr/SongSZSS023}
G.~Song, Q.~Sun, L.~Zhang, R.~Su, J.~Shi, and Y.~He, ``{OPE-SR:} orthogonal position encoding for designing a parameter-free upsampling module in arbitrary-scale image super-resolution,'' in \emph{CVPR}, 2023.

\bibitem{DBLP:conf/cvpr/LeeJ22}
J.~Lee and K.~H. Jin, ``Local texture estimator for implicit representation function,'' in \emph{CVPR}, 2022.

\bibitem{DBLP:conf/cvpr/Cao0XLNP0ZTG23}
J.~Cao, Q.~Wang, Y.~Xian, Y.~Li, B.~Ni, Z.~Pi, K.~Zhang, Y.~Zhang, R.~Timofte, and L.~V. Gool, ``Ciaosr: Continuous implicit attention-in-attention network for arbitrary-scale image super-resolution,'' in \emph{CVPR}, 2023.

\bibitem{DBLP:conf/cvpr/ChenXHTKL23}
H.~Chen, Y.~Xu, M.~Hong, Y.~Tsai, H.~Kuo, and C.~Lee, ``Cascaded local implicit transformer for arbitrary-scale super-resolution,'' in \emph{CVPR}, 2023.

\bibitem{DBLP:conf/cvpr/YaoTLTCL23}
J.~Yao, L.~Tsao, Y.~Lo, R.~Tseng, C.~Chang, and C.~Lee, ``Local implicit normalizing flow for arbitrary-scale image super-resolution,'' in \emph{CVPR}, 2023.

\bibitem{DBLP:conf/cvpr/GaoLZXLLLZ023}
S.~Gao, X.~Liu, B.~Zeng, S.~Xu, Y.~Li, X.~Luo, J.~Liu, X.~Zhen, and B.~Zhang, ``Implicit diffusion models for continuous super-resolution,'' in \emph{CVPR}, 2023.

\bibitem{DBLP:conf/eccv/LeeCJ22}
J.~Lee, K.~P. Choi, and K.~H. Jin, ``Learning local implicit fourier representation for image warping,'' in \emph{ECCV}, 2022.

\bibitem{Xiao_2024_CVPR}
J.~Xiao, Z.~Lyu, C.~Zhang, Y.~Ju, C.~Shui, and K.-M. Lam, ``Towards progressive multi-frequency representation for image warping,'' in \emph{CVPR}, 2024.

\bibitem{DBLP:conf/cvpr/PakLJ23}
B.~Pak, J.~Lee, and K.~H. Jin, ``B-spline texture coefficients estimator for screen content image super-resolution,'' in \emph{CVPR}, 2023.

\bibitem{DBLP:journals/corr/abs-2112-09318}
B.~Wronski, ``Procedural kernel networks,'' \emph{arxiv}, vol. 2112.09318, 2021.

\bibitem{DBLP:journals/pami/KimLLSLB12}
S.~J. Kim, H.~T. Lin, Z.~Lu, S.~S{\"{u}}sstrunk, S.~Lin, and M.~S. Brown, ``A new in-camera imaging model for color computer vision and its application,'' \emph{{IEEE} Trans. Pattern Anal. Mach. Intell.}, vol.~34, no.~12, pp. 2289--2302, 2012.

\bibitem{DBLP:journals/pami/ZengCLCZ22}
H.~Zeng, J.~Cai, L.~Li, Z.~Cao, and L.~Zhang, ``Learning image-adaptive 3d lookup tables for high performance photo enhancement in real-time,'' \emph{{IEEE} Trans. Pattern Anal. Mach. Intell.}, vol.~44, no.~4, pp. 2058--2073, 2022.

\bibitem{DBLP:conf/iccv/LiuDLSWW23}
G.~Liu, Y.~Ding, M.~Li, M.~Sun, X.~Wen, and B.~Wang, ``Reconstructed convolution module based look-up tables for efficient image super-resolution,'' in \emph{ICCV}, 2023.

\bibitem{DBLP:journals/tog/YuT13}
J.~Yu and G.~Turk, ``Reconstructing surfaces of particle-based fluids using anisotropic kernels,'' \emph{{ACM} Trans. Graph.}, vol.~32, no.~1, pp. 5:1--5:12, 2013.

\bibitem{DBLP:conf/mm/HuiGYW19}
Z.~Hui, X.~Gao, Y.~Yang, and X.~Wang, ``Lightweight image super-resolution with information multi-distillation network,'' in \emph{ACM Int. Conf. Multimedia}, 2019.

\bibitem{DBLP:conf/cvpr/AgustssonT17}
E.~Agustsson and R.~Timofte, ``{NTIRE} 2017 challenge on single image super-resolution: Dataset and study,'' in \emph{CVPR Workshops}, 2017.

\bibitem{DBLP:conf/iccv/MartinFTM01}
D.~R. Martin, C.~C. Fowlkes, D.~Tal, and J.~Malik, ``A database of human segmented natural images and its application to evaluating segmentation algorithms and measuring ecological statistics,'' in \emph{ICCV}, 2001.

\bibitem{DBLP:conf/cvpr/HuangSA15}
J.~Huang, A.~Singh, and N.~Ahuja, ``Single image super-resolution from transformed self-exemplars,'' in \emph{CVPR}, 2015.

\bibitem{DBLP:journals/mta/MatsuiIAFOYA17}
Y.~Matsui, K.~Ito, Y.~Aramaki, A.~Fujimoto, T.~Ogawa, T.~Yamasaki, and K.~Aizawa, ``Sketch-based manga retrieval using manga109 dataset,'' \emph{Multim. Tools Appl.}, vol.~76, no.~20, pp. 21\,811--21\,838, 2017.

\bibitem{DBLP:journals/tip/WangBSS04}
Z.~Wang, A.~C. Bovik, H.~R. Sheikh, and E.~P. Simoncelli, ``Image quality assessment: from error visibility to structural similarity,'' \emph{{IEEE} Trans. Image Process.}, vol.~13, no.~4, pp. 600--612, 2004.

\bibitem{DBLP:conf/cvpr/ZhangIESW18}
R.~Zhang, P.~Isola, A.~A. Efros, E.~Shechtman, and O.~Wang, ``The unreasonable effectiveness of deep features as a perceptual metric,'' in \emph{CVPR}, 2018.

\bibitem{DBLP:journals/corr/KingmaB14}
D.~P. Kingma and J.~Ba, ``Adam: {A} method for stochastic optimization,'' in \emph{ICLR}, 2015.

\bibitem{DBLP:conf/iclr/LoshchilovH17}
I.~Loshchilov and F.~Hutter, ``{SGDR:} stochastic gradient descent with warm restarts,'' in \emph{ICLR}, 2017.

\bibitem{10.2307/2030559}
J.~M. Borwein and P.~B. Borwein, ``On the complexity of familiar functions and numbers,'' \emph{SIAM Review}, vol.~30, no.~4, pp. 589--601, 1988.

\bibitem{DBLP:conf/cvpr/ZhangTKZ018}
Y.~Zhang, Y.~Tian, Y.~Kong, B.~Zhong, and Y.~Fu, ``Residual dense network for image super-resolution,'' in \emph{CVPR}, 2018.

\bibitem{DBLP:conf/dagm/AurichW95}
V.~Aurich and J.~Weule, ``Non-linear gaussian filters performing edge preserving diffusion,'' in \emph{Mustererkennung}, 1995.

\bibitem{DBLP:conf/iccv/TomasiM98}
C.~Tomasi and R.~Manduchi, ``Bilateral filtering for gray and color images,'' in \emph{ICCV}, 1998.

\bibitem{DBLP:journals/ftcgv/KornprobstTD09}
P.~Kornprobst, J.~Tumblin, and F.~Durand, ``Bilateral filtering: Theory and applications,'' \emph{Found. Trends Comput. Graph. Vis.}, vol.~4, no.~1, pp. 1--74, 2009.

\bibitem{DBLP:journals/tog/KerblKLD23}
B.~Kerbl, G.~Kopanas, T.~Leimk{\"{u}}hler, and G.~Drettakis, ``3d gaussian splatting for real-time radiance field rendering,'' \emph{{ACM} Trans. Graph.}, vol.~42, no.~4, pp. 139:1--139:14, 2023.

\end{thebibliography}
